\documentclass[10pt,journal,letterpaper,compsoc]{IEEEtran}

\usepackage{times}
\usepackage{epsfig}
\usepackage{enumerate}
\usepackage{graphicx}
\usepackage{amsmath}
\usepackage{amssymb}
\usepackage{diagbox}
\usepackage{multirow}
\usepackage{subfigure}
\usepackage{color}
\usepackage{verbatim}
\usepackage{array}
\usepackage{url}
\usepackage{bm}
\usepackage[pagebackref=true,breaklinks=true,letterpaper=true,colorlinks,bookmarks=false]{hyperref}
\newcommand{\tabincell}[2]{\begin{tabular}{@{}#1@{}}#2\end{tabular}}

\begin{document}
\hyphenpenalty=1000
%
\title{Face Alignment in Full Pose Range: \\A 3D Total Solution}
%
%
%

\author{Xiangyu Zhu, Xiaoming Liu,~\IEEEmembership{Member,~IEEE,} Zhen Lei,~\IEEEmembership{Senior Member,~IEEE,} and Stan Z. Li,~\IEEEmembership{Fellow,~IEEE}

\IEEEcompsocitemizethanks{\IEEEcompsocthanksitem X. Zhu, Z. Lei and S. Li are with Center for Biometrics and Security Research \& National Laboratory of Pattern Recognition, Institute of Automation,
Chinese Academy of Sciences, 95 Zhongguancun Donglu, Beijing 100190,
China. \protect Email: \{xiangyu.zhu,zlei,szli\}@nlpr.ia.ac.cn.

\IEEEcompsocthanksitem X. Liu is with the Department of Computer Science and Engineering, Michigan State University, East Lansing, MI 48824, USA. \protect Email: liuxm@msu.edu.}
\thanks{}}

\IEEEcompsoctitleabstractindextext{%
\begin{abstract}
\renewcommand{\raggedright}{\leftskip=0pt \rightskip=0pt plus 0cm}
Face alignment, which fits a face model to an image and extracts the semantic meanings of facial pixels, has been an important topic in the computer vision community. However, most algorithms are designed for faces in small to medium poses (yaw angle is smaller than $45^{\circ}$), which lack the ability to align faces in large poses up to $90^{\circ}$. The challenges are three-fold. Firstly, the commonly used landmark face model assumes that all the landmarks are visible and is therefore not suitable for large poses. Secondly, the face appearance varies more drastically across large poses, from the frontal view to the profile view. Thirdly, labelling landmarks in large poses is extremely challenging since the invisible landmarks have to be guessed. In this paper, we propose to tackle these three challenges in an new alignment framework termed 3D Dense Face Alignment (3DDFA), in which a dense 3D Morphable Model (3DMM) is fitted to the image via Cascaded Convolutional Neural Networks. We also utilize 3D information to synthesize face images in profile views to provide abundant samples for training. Experiments on the challenging AFLW database show that the proposed approach achieves significant improvements over the state-of-the-art methods.
\raggedright
\end{abstract}

\begin{IEEEkeywords}
Face Alignment, 3D Morphable Model, Convolutional Neural Network, Cascaded Regression
\end{IEEEkeywords}}

\maketitle

\section{Introduction}\label{sec-introduction}
Face alignment is the process of moving and deforming a face model to an image, so as to extract the semantic meanings of facial pixels. 
It is an essential preprocessing step for many face analysis tasks, e.g. recognition~\cite{Taigman-CVPR-2013}, animation~\cite{Cao-SIG-2013}, tracking~\cite{xiong2015global}, attributes classification~\cite{bettadapura2012face} and image restoration~\cite{yang2013structured}.
Traditionally, face alignment is approached as a landmark detection problem that aims to locate a sparse set of facial fiducial points, some of which include ``eye corner'', ``nose tip'' and ``chin center''. In the past two decades, a number of effective frameworks have been proposed such as ASM~\cite{cootes1995active}, AAM~\cite{Cootes-ECCV-98} and CLM~\cite{Cristinacce-BMVC-06}. Recently, with the introduction of Cascaded Regression~\cite{Dollar-CVPR-10,Cao-CVPR-12,Xiong-CVPR-13} and Convolutional Neural Networks~\cite{sun2013deep,zhang2014facial}, face alignment has observed significant improvements in accuracy.
However, most of the existing methods are designed for medium poses, under the assumptions that the yaw angle is smaller than $45^{\circ}$ and all the landmarks are visible. When the range of yaw angle is extended up to $90^{\circ}$, significant challenges emerge. These challenges can be differentiated in three main ways:

\begin{figure}[!htb]
  \centering
  \includegraphics[width=0.45\textwidth]{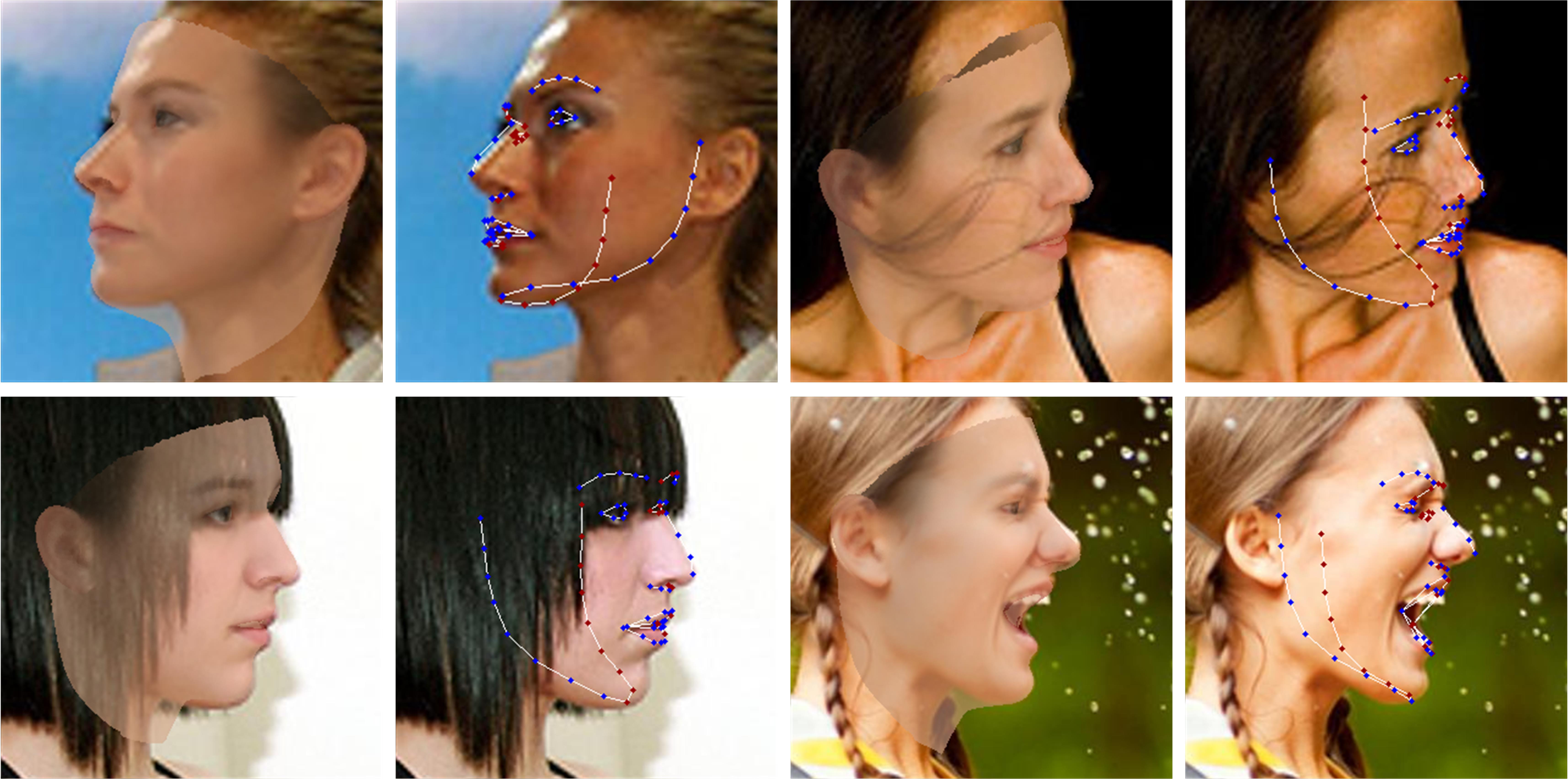}
  \caption{Fitting results of 3DDFA (the blue/red points indicate visible/invisible landmarks). For each pair of the four results, on the left is the rendering of the fitted 3D face with the mean texture, which is made transparent to demonstrate the fitting accuracy. On the right is the landmarks overlayed on the fitted 3D face model.}
  \label{fig-demo}
\end{figure}

\textbf{Modelling}: Landmark shape model~\cite{cootes1995active} implicitly assumes that each landmark can be robustly detected by its distinctive visual patterns. However, when faces deviate from the frontal view, some landmarks become invisible due to self-occlusion~\cite{zhou2005bayesian}. In medium poses, this problem can be addressed by changing the semantic positions of face contour landmarks to the silhouette, which is termed landmark marching~\cite{zhu2015high}. However, in large poses where half of face is occluded, some landmarks are inevitably invisible and show no detectable appearance. In turn, landmarks can lose their semantic meanings, which may cause the shape model to fail.

\textbf{Fitting}: Another challenge in full-pose face alignment is derived from the dramatic appearance variations from front to profile. Cascaded Linear Regression~\cite{Xiong-CVPR-13} and traditional nonlinear models~\cite{zhang2014coarse,Cao-CVPR-12} are not flexible enough to cover these complex variations in a unified way. Another framework demonstrates more flexibility by adopting different landmark and fitting models for differing view categories~\cite{zhou2005bayesian,yu2013pose,zhu2012face}. Unfortunately, since the nature of this framework must test every view, computational cost is likely to significantly increase. More recently, Convolutional Neural Network (CNN) based methods have demonstrated improved performance over traditional methods in many applications. For effective large-pose face alignment, CNN should be combined with the Cascaded Regression framework. However, most existing methods adopt a single network to complete fitting~\cite{zhang2014facial}, which limits its performance.

\textbf{Training Data}: Labelled data is the basis for any supervised learning based algorithms. However, manual labelling of landmarks on large-pose faces is very tedious since the occluded landmarks have to be ``guessed'' which is impossible for most of people. As a result, almost all the public face alignment databases such as AFW~\cite{zhu2012face}, LFPW~\cite{jaiswal2013guided}, HELEN~\cite{le2012interactive} and IBUG~\cite{sagonas2013semi} are collected in medium poses. Few large-pose databases such as AFLW~\cite{kostinger2011annotated} only contain visible landmarks, which could be ambiguous in invisible landmarks, makes it hard to train a unified face alignment model.

In this paper, we aim to solve the problem of face alignment in full pose range, where the yaw angle is allowed to vary between $\pm 90^{\circ}$. We believe that face alignment is not barely a 2D problem since self-occlusion and large appearance variations are caused by the face rotation in the 3D space, which can be conveniently addressed by incorporating 3D information.
More specifically, we improve the face model from 2D sparse landmarks to a dense 3D Morphable Model (3DMM)~\cite{Blanz-PAMI-03} and consider face alignment as a 3DMM fitting task. The optimization concept therein will change accordingly from landmark positions to pose (scale, rotation and translation) and morphing (shape and expression) parameters. We call this novel face alignment framework 3D Dense Face Alignment (\textbf{3DDFA}).
To realize 3DDFA, we propose to combine two achievements in recent years, namely, Cascaded Regression and the Convolutional Neural Network (CNN). This combination requires the introduction of a new \textbf{input feature} which fulfills the ``cascade manner'' and ``convolution manner'' simultaneously (see Sec.~\ref{sec-feature-design}) and a new \textbf{cost function} which can model the priority of 3DMM parameters (see Sec.~\ref{sec-cost-function}).
Besides to provide enough data for training, we find that given a face image and its corresponding 3D model, it is possible to rotate the image out of plane with high fidelity. This rotation enables the synthesis of a large number of training samples in large poses.

In general, we propose a novel face alignment framework to address the three challenges of modelling, fitting and training data in large poses. The main contributions of the paper are summarized as follows:
\begin{enumerate}
\item To address the self-occlusion challenge, we assert that in large poses, fitting a 3DMM is more suitable than detecting 2D landmarks. The visibility estimated from 3DMM enables us to only fit the vertices with detected image patterns. The landmarks, if needed, can be sampled from the fitted 3D face afterwards. See the samples in Fig.~\ref{fig-demo}.
\item To handle appearance variations across large poses, we propose a novel Cascaded Convolutional Neural Network as the regressor, in which two specially designed input features called Projected Normalized Coordinate Code (PNCC) and Pose Adaptive Feature (PAF) are introduced to connect CNNs in a cascade manner. Besides, a novel cost function named Optimized Weighted Parameter Distance Cost (OWPDC) is proposed to formulate the priority of 3DMM parameters during training.
\item To enable the training of the 3DDFA, we construct a face database consisting of pairs of 2D face images and 3D face models. We further elucidate a face profiling method to synthesize $60k+$ training samples across large poses. The synthesized samples well simulate the face appearances in large poses and boost the performance of both previous and the proposed face alignment approaches.
\end{enumerate}

This paper is an extension of our previous work~\cite{zhu2016face} the following four aspects:
1) Traditional 3DMM uses Euler angles to represent the 3D rotation, which shows ambiguity when the yaw angle reaches $90^{\circ}$. In this paper, quaternions are used instead as the rotation formulation to eliminate the ambiguity. 2) A new input feature called Pose Adaptive Feature (PAF) is utilized to remedy the drawbacks of PNCC to further boost the performance. 3) We improve the cost function in~\cite{zhu2016face} through the OWPDC which not only formulates the importance but also the priority of 3DMM parameters during training. 4) Additional experiments are conducted to better analyze the motivation behind the design of the input features and the cost function.

\section{Related Works}
Face alignment can be summarized as \textbf{fitting} a \textbf{face model} to an image. As such, there are two basic problems involved with this task: how to model the face shape and how to estimate the model parameters. In this section, we motivate our approach by discussing related works with respect to these two problems.

\subsection{Face Model}
Traditionally, face shape is represented by a sparse set of 2D facial fiducial points. Cootes et al.~\cite{cootes1995active,Cootes-ECCV-98} show that shape variations can be modeled with subspace analysis such as Principal Components Analysis (PCA). Although, this \textbf{2D-subspace model} can only cope with shape variations from a narrow range of face poses, since the non-linear out-of-plane rotation cannot be well represented with the linear subspace. To deal with the pose variations, some modifications like Kernel PCA~\cite{romdhani1999a} and Bayesian Mixture Model~\cite{zhou2005bayesian} are proposed to introduce non-linearity into the subspace models. Recently, Cao et al.~\cite{Cao-CVPR-12} propose to abandon any explicit shape constraints and directly use landmark coordinates as the shape model, which called 2D Non-Parametric Model (\textbf{2D-NPM}). 2D-NPM considerably improves the flexibility of the shape model at the cost of losing any shape priors and increasing the difficulty of model fitting. Besides 2D shape model, Blanz et al.~\cite{Blanz-SIG-99,Blanz-PAMI-03} propose the 3D Morphable Model (\textbf{3DMM}) which applies PCA on a set of 3D face scans. By incorporating 3D information, 3DMM disentangles the non-linear out-of-plane transformation from the PCA subspace. The remaining shape and expression variations have shown high linearity~\cite{Blanz-PAMI-03,Cao-SIG-2013}, which can be well modeled with PCA. Compared with 2D models, 3DMM separates rigid (pose) and non-rigid (shape and expression) transformations, enabling it to cover diverse shape variations and keep shape prior at the same time. Additionally, points visibility can be easily estimated by 3DMM~\cite{zhu2016face}, which can provide important clues to handle self-occlusion in profile views.

\subsection{Model Fitting}
Most fitting methods can be divided into two categories: the template fitting based~\cite{Cootes-ECCV-98,Cristinacce-PR-08} and regression based~\cite{cootes1998comparative,Dollar-CVPR-10,Xiong-CVPR-13,trigeorgis2016mnemonic}. The template fitting methods always maintain a face appearance model to fit images. For example, Active Appearance Model (AAM)~\cite{Cootes-ECCV-98} and Analysis-by-Synthesis 3DMM Fitting~\cite{Blanz-PAMI-03} simulate the process of face image generation and
achieve alignment by minimizing the difference between the model appearance and the input image. Active Shape Model (ASM)~\cite{cootes1995active} and Constrained Local Model (CLM)~\cite{Cristinacce-BMVC-06,Saragih-IJCV-10} build a template model for each landmark and use a PCA shape model to constrain the fitting results. TSPM~\cite{zhu2012face} and CDM~\cite{yu2013pose} employ part based model and DPM-like~\cite{felzenszwalb2010object} method to align faces. Generally, the performance of template fitting methods depends on whether the image patterns reside within the variations described by the face appearance model. Therefore, it shows limited robustness in unconstrained environment where appearance variations are too wide and complicated.

Regression based methods estimate model parameters by regressing image features. For example, Hou et al.~\cite{hou2001direct} and Saragih et al.~\cite{saragih2007nonlinear} perform regression between texture residuals and parameter updates to fit AAM. Valstar et al.~\cite{Valstar-CVPR-10} locate landmark positions by mapping the landmark related local patches with support vector regression. Recently, Cascaded Regression~\cite{Dollar-CVPR-10} has been proposed and becomes most popular in face alignment community~\cite{Cao-CVPR-12,Xiong-CVPR-13,ren2014face,zhu2015face}, which can be summarized in Eqn.~\ref{equ-cascaded-regression}:
\begin{gather}\label{equ-cascaded-regression}
  \mathbf{p}^{k+1} = \mathbf{p}^{k} + \emph{Reg}^{k} (\emph{Fea}(\mathbf{I},\mathbf{p}^{k})).
\end{gather}
where the shape parameter $\mathbf{p}^{k}$ at the $k$th iteration is updated by conducting regression $\emph{Reg}^{k}$ on the shape indexed feature $\emph{Fea}$, which should depend on both the image $\mathbf{I}$ and the current parameter $\mathbf{p}^{k}$. The regression $\emph{Reg}^{k}$ shows an important ``feedback'' property that its input feature $\emph{Fea}(\mathbf{I},\mathbf{p})$ can be updated by its output since after each iteration $\mathbf{p}$ is updated. With this property an array of weak regressors can be cascaded to reduce the alignment error progressively.

Besides Cascaded Regression, another breakthrough is the introduction of Convolutional Neural Network (CNN), which formulates face alignment as a regression from raw pixels to landmarks positions. For example, Sun et al.~\cite{sun2013deep} propose to use the CNN to locate landmarks in two stages, first the full set of landmarks are located with a global CNN and then each landmark is refined with a sub-network on its local patch. With one CNN for each landmark, the complexity of the method highly depends on the number of landmarks. Zhang et al.~\cite{zhang2014facial} combine face alignment with attribute analysis through multi-task CNN to boost the performance of both tasks. Wu et al.~\cite{wu2015facial} cluster face appearances with mid-level CNN features and deal with each cluster with an independent regressor. Jourabloo et al.~\cite{jourabloo2016large} arrange the local landmark patches into a large 2D map as the CNN input to regress model parameters. Trigeorgis et al.~\cite{trigeorgis2016mnemonic} convolve the landmark local patch as the shape index feature and conduct linear regression to locate landmarks.

\subsection{Large Pose Face Alignment}
Despite the great achievements in face alignment, most of the state-of-the-art methods lack the flexibility in large-pose scenarios, since they need to build the challenging relationship between the landmark displacement and landmark related image features, where the latter may be self-occluded. In 2D methods, a common solution is the multi-view framework which uses different landmark configurations for different views. It has been applied in AAM~\cite{cootes2002view}, DAM~\cite{li2002multi} and DPM~\cite{zhu2012face,yu2013pose} to align faces with different shape models, among which the one having the highest possibility is chosen as the final result. However, since every view has to be tested, the computational cost is always high. Another method is explicitly estimating the visibility of landmarks and shrink the contribution of occluded features~\cite{zhou2005bayesian,gross2006active,burgos2013robust}. Nevertheless, occlusion estimation is itself a challenging task and handling varying dimensional feature is still an ill-posed problem.

Different from 2D methods, 3D face alignment~\cite{gu20063D} aims to fit a 3DMM~\cite{Blanz-PAMI-03} to a 2D image. By incorporating 3D information, 3DMM can inherently provide the visibility of each model point without any additional estimation, making it possible to deal with the self-occluded points. The original 3DMM fitting method~\cite{Blanz-PAMI-03} fits the 3D model by minimizing the pixel-wise difference between image and rendered face model. Since only the visible model vertices are fitted, it is the first method to cover arbitrary poses~\cite{Blanz-PAMI-03,Romdhani-CVPR-05}, but it suffers from the one-minute-per-image computational cost. Recently, regression based 3DMM fitting, which estimates the model parameters by regressing the features at projected 3D landmarks~\cite{yu2013pose,jourabloo2015pose,Cao-2014-SIG,jeni2015dense,jourabloo2016large,jourabloo2017pose,jourabloo2017alignment}, has looked to improve the efficiency. Although these methods face two major challenges. First the projected 3D landmarks may be self-occluded and lose their image patterns, making the features no longer pose invariant. Second, parameters of 3DMM have different priorities during fitting, despite that existing regression based methods treat them equally~\cite{Cao-CVPR-12}. As a result, directly minimizing the parameter error may be sub-optimal, because smaller parameter errors are not necessarily equivalent to smaller alignment errors. This problem will be further discussed in Sec.~\ref{sec-cost-function}.
A relevant but distinct task is 3D face reconstruction~\cite{Aldrian-PAMI-13,zhu2015high,hassner2013viewing,roth2016adaptive}, which recovers a 3D face from given 2D landmarks. Interestingly, 2D/3D face alignment results can be mutually transformed, where 3D to 2D is made by sampling landmark vertices and 2D to 3D is made by 3D face reconstruction.

In this work, we propose a framework to combine three major achievements---3DMM, Cascaded Regression and CNN---to solve the large-pose face alignment problem.

\begin{figure*}
  \centering
  \includegraphics[width=0.90\textwidth]{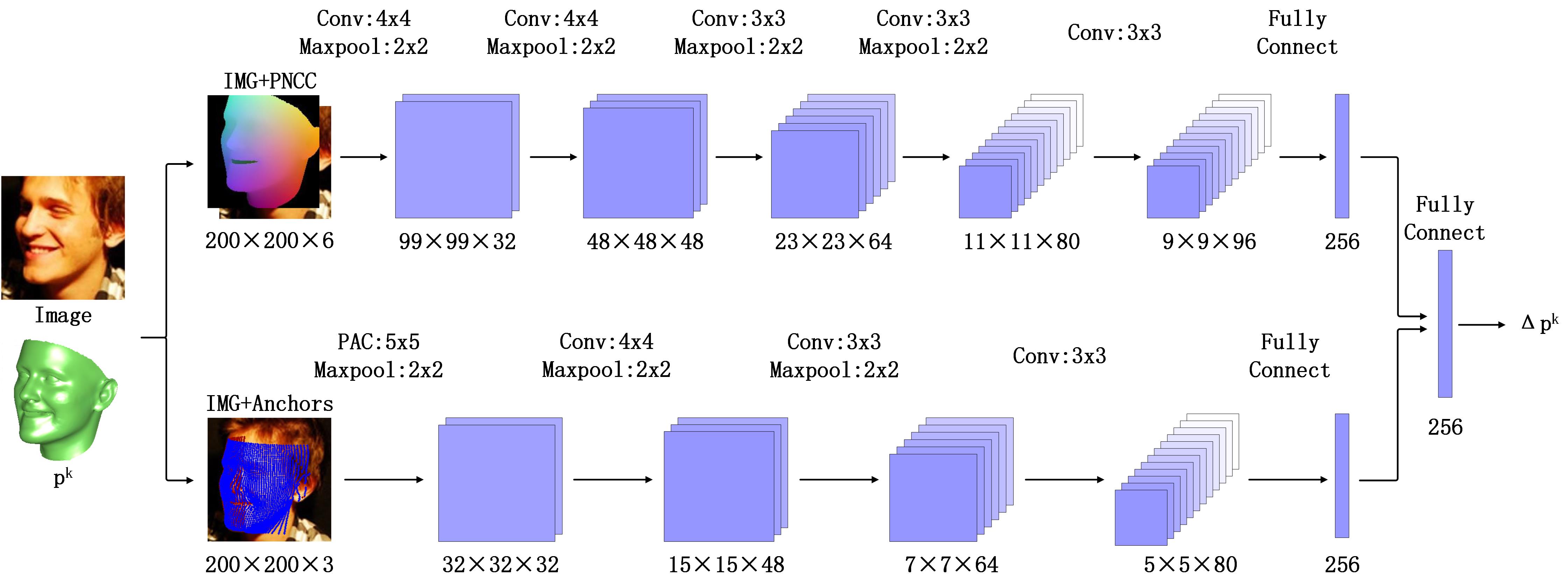}
  \caption{An overview of the two-stream network in 3DDFA. With an intermediate parameter $\mathbf{p}^{k}$, in the first stream we construct a novel Projected Normalized Coordinate Code (PNCC), which is stacked with the input image and sent to the CNN. In the second stream, we get some feature anchors with consistent semantics and conduct Pose Adaptive Convolution (PAC) on them. The outputs of the two streams are merged with an additional fully connected layer to predict the parameter update $\Delta \mathbf{p}^{k}$.}
  \label{fig-overview}
\end{figure*}

\section{3D Dense Face Alignment (3DDFA)}
In this section, we introduce how to combine Cascaded Regression and CNNs to realize 3DDFA. By applying a CNN as the regressor in Eqn.~\ref{equ-cascaded-regression}, Cascaded CNN can be formulated as:
\begin{equation}\label{equ-cascaded-CNN}
  \mathbf{p}^{k+1} = \mathbf{p}^{k} + \emph{Net}^{k} (\emph{Fea}(\mathbf{I},\mathbf{p}^{k})).
\end{equation}
There are four components in this framework: the regression objective $\mathbf{p}$ (Sec.~\ref{sec-3D-Morphable-Model}), the image features $\emph{Fea}$ (Sec.~\ref{sec-feature-design}), the CNN structure $\emph{Net}$ (Sec.~\ref{sec-network-structure}) and the cost function to train the framework (Sec.~\ref{sec-cost-function}).

\subsection{3D Morphable Model}\label{sec-3D-Morphable-Model}
Blanz et al.~\cite{Blanz-PAMI-03} propose the 3D Morphable Model (3DMM) to describe the 3D face space with PCA:
\begin{equation}\label{equ-tensor}
  \mathbf{S}=\mathbf{\overline{S}} + \mathbf{A}_{id}\bm{\alpha}_{id} + \mathbf{A}_{exp}\bm{\alpha}_{exp},
\end{equation}
where $\mathbf{S}$ is a 3D face, $\mathbf{\overline{S}}$ is the mean shape, $\mathbf{A}_{id}$ is the principle axes trained on the 3D face scans with neutral expression and $\bm{\alpha}_{id}$ is the shape parameter, $\mathbf{A}_{exp}$ is the principle axes trained on the offsets between expression scans and neutral scans and $\bm{\alpha}_{exp}$ is the expression parameter. In this work, the $\mathbf{A}_{id}$ and $\mathbf{A}_{exp}$ come from BFM~\cite{Paysan-AVSS-09} and FaceWarehouse~\cite{Cao-2013-Facewarehouse} respectively. After the 3D face is constructed, it can be projected onto the image plane with scale orthographic projection:
\begin{equation}\label{equ-projection}
  V(\mathbf{p}) = f*\mathbf{Pr} * \mathbf{R}*(\mathbf{\overline{S}} + \mathbf{A}_{id}\bm{\alpha}_{id} + \mathbf{A}_{exp}\bm{\alpha}_{exp}) +\mathbf{t}_{2d},
\end{equation}
where $V(\mathbf{p})$ is the model construction and projection function, leading to the 2D positions of model vertices, $f$ is the scale factor, $\mathbf{Pr}$ is the orthographic projection matrix
$\left(
\begin{array}{ccc}
1 & 0 & 0 \\
0 & 1 & 0 \\
\end{array}
\right)
$
, $\mathbf{R}$ is the rotation matrix and $\mathbf{t}_{2d}$ is the translation vector. The collection of all the model parameters is $\mathbf{p}=[f,\mathbf{R},\mathbf{t}_{2d},\bm{\alpha}_{id},\bm{\alpha}_{exp}]^\mathrm{T}$.

\subsubsection{Rotation Formulation}
\begin{figure}[!htb]
  \centering
  \includegraphics[width=0.40\textwidth]{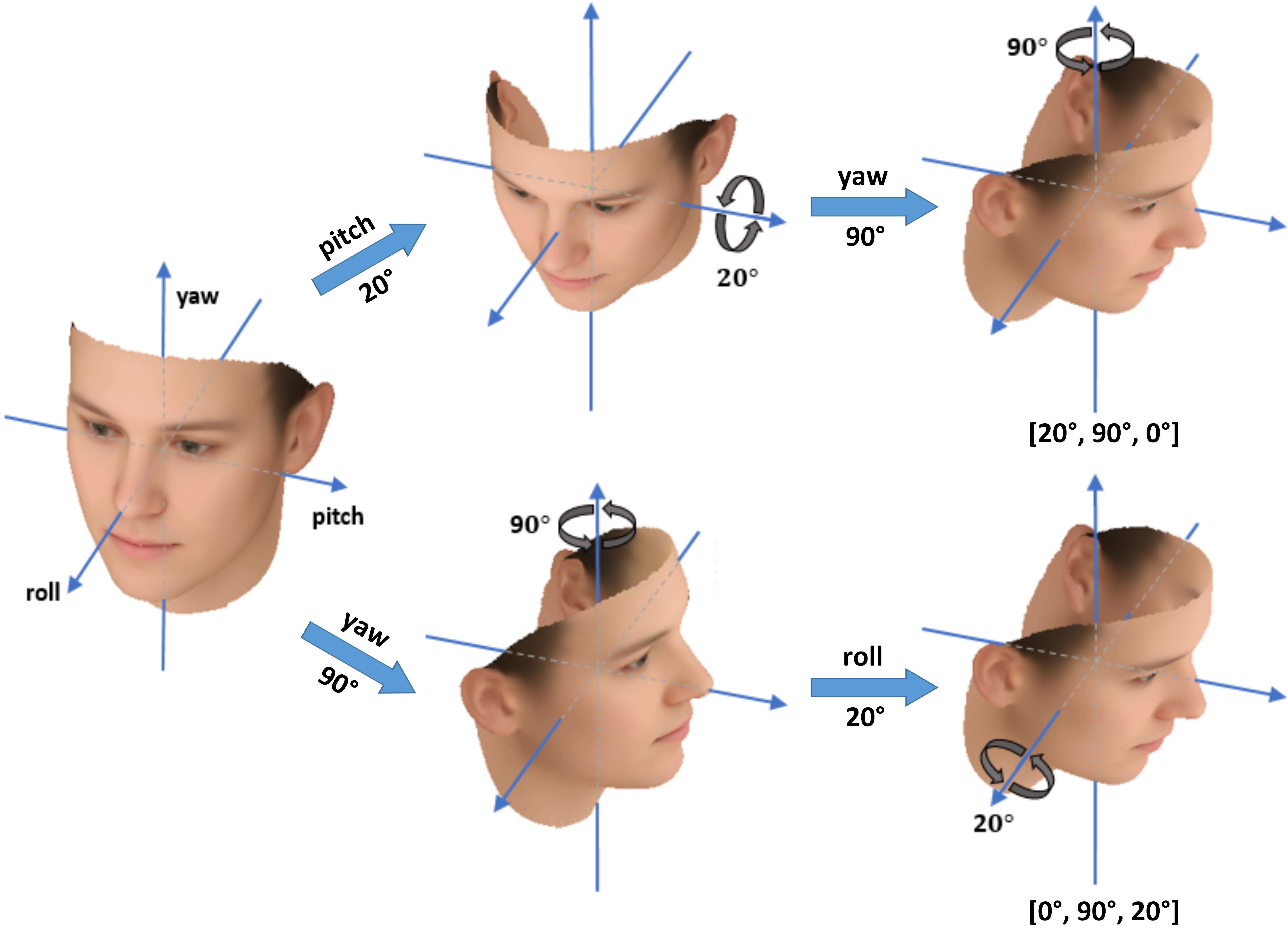}
  \caption{An example of gimbal lock. We assume the rotation sequence is from $pitch$ to $yaw$ to $roll$. In the first row, the face is firstly rotated $20^{\circ}$ around the $pitch$ axis and then $90^{\circ}$ around the $yaw$ axis, whose Euler angles are $[20^{\circ}, 90^{\circ}, 0^{\circ}]$. In the second row, the face is firstly rotated $90^{\circ}$ around the $yaw$ axis and then $20^{\circ}$ around the $roll$ axis, whose Euler angles are $[0^{\circ}, 90^{\circ}, 20^{\circ}]$. However the two different Euler angles correspond to the same rotation matrix, generating the profile view of a nodding face.}
  \label{fig-gimbal-lock}
\end{figure}

Face rotation is traditionally formulated with the Euler angles~\cite{murphy2009head} including $pitch$, $yaw$ and $roll$. However, when faces are close to the profile view, there is ambiguity in Euler angles termed gimbal lock~\cite{lepetit2005monocular}, see Fig.~\ref{fig-gimbal-lock} as a example.

The ambiguity in Euler angles will confuse the regressor and affect the fitting performance. Therefore we adopt a four dimensional unit quaternion~\cite{lepetit2005monocular} $[q_{0}, q_{1}, q_{2}, q_{3}]$ instead of the Euler angles to formulate the rotation. The corresponding rotation matrix is:
\begin{small}
\begin{gather}\label{equ-rotation-quat}
\mathbf{R} =
\begin{bmatrix}
q_{0}^{2}+q_{1}^{2}-q_{2}^{2}-q_{3}^{2} & 2(q_{1}q_{2}+q_{0}q_{3}) & 2(q_{1}q_{3}-q_{0}q_{2}) \\
2(q_{1}q_{2}-q_{0}q_{3}) & q_{0}^{2}-q_{1}^{2}+q_{2}^{2}-q_{3}^{2} & 2(q_{0}q_{1}+q_{2}q_{3}) \\
2(q_{0}q_{2}+q_{1}q_{3}) & 2(q_{2}q_{3}-q_{0}q_{1}) & q_{0}^{2}-q_{1}^{2}-q_{2}^{2}+q_{3}^{2} \notag
\end{bmatrix}
\end{gather}
\end{small}
In our implementation, we merge the scale parameter $f$ into $[q_{0}, q_{1}, q_{2}, q_{3}]$ through dividing the quaternion by $\sqrt{f}$ and do not constrain the quaternion to be unit. As a result, the fitting objective will be $\mathbf{p}=[q_{0}, q_{1}, q_{2}, q_{3},\mathbf{t}_{2d},\bm{\alpha}_{id},\bm{\alpha}_{exp}]^\mathrm{T}$.

\subsection{Feature Design}\label{sec-feature-design}
As the conjunction point of Cascaded Regression and CNN, the input feature should fulfill the requirements from both frameworks, which can be summarized as the following three aspects: Firstly, the \textbf{convolvable property} requires that the convolution operation on the input feature should make sense. As the CNN input, the feature should be a smooth
2D map reflecting the accuracy of current fitting. Secondly, to enable the cascade manner, the \textbf{feedback property} requires the input feature to depend on the CNN output~\cite{Dollar-CVPR-10}. Finally, to guarantee the cascade to converge at the ground truth parameter, the \textbf{convergence property} requires the input feature to be discriminative when the fitting is complete.

Besides the three requirements, we find that the input features of face alignment can be divided into two categories. The first category is the image-view feature, where the original image is directly sent to the regressor. For example, \cite{sun2013deep,zhang2014facial,wu2015facial} use the input image as the CNN input and \cite{liang2015unconstrained,carreira2015human} stack the image with a landmark response map as the input. These kind of features does not lose any information provided by the image but require the regressor to cover any face appearances. The second category is the model-view feature, where image pixels are rearranged based on the model condition. For example, AAM~\cite{Cootes-ECCV-98} warps the face image to the mean shape and SDM~\cite{Xiong-CVPR-13} extract SIFT features at landmark locations. This kind of features aligns the face appearance with current fitting, which simplifies the alignment task progressively during optimization. However, they do not cover the pixels beyond the face model, leading to a bad description of context. As such, fitting with model-view features is easily trapped in local minima~\cite{zhu2015face}. In this paper, we propose a model-view feature called Pose Adaptive Feature (PAF) and a image-view feature called Projected Normalized Coordinate Code (PNCC). We further demonstrate that optimal results can be achieved by combining both features.

\subsubsection{Pose Adaptive Convolution}
\begin{figure}[!htb]
  \centering
  \subfigure[]{\label{fig-PAF-a}
  \includegraphics[width=0.11\textwidth]{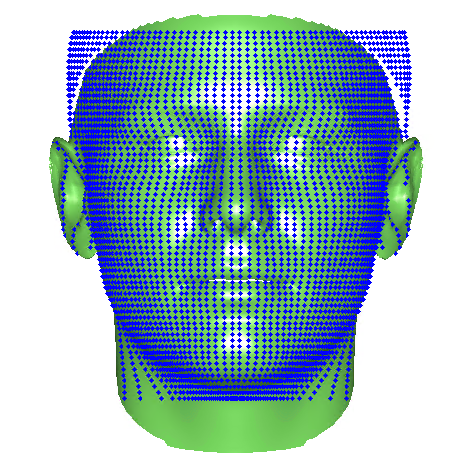}}
  \subfigure[]{\label{fig-PAF-b}
  \includegraphics[width=0.11\textwidth]{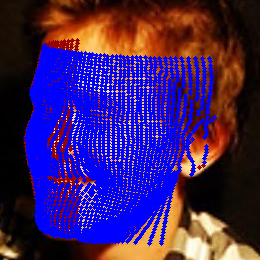}}
  \subfigure[]{\label{fig-PAF-c}
  \includegraphics[width=0.11\textwidth]{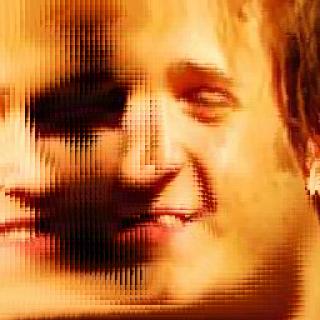}}
  \subfigure[]{\label{fig-PAF-d}
  \includegraphics[width=0.11\textwidth]{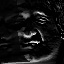}}
  \caption{Pose Adaptive Convolution (PAC): (a) The $64 \times 64$ feature anchors on the 3D face model. (b) The projected feature anchors $V(\mathbf{p})_{anchor}$ (the blue/red ones indicate visible/invisible anchors). (c) The feature patch map concatenated by the patches cropped at $V(\mathbf{p})_{anchor}$. (d) Conducting convolution, whose stride and filter size are the same with the patch size, on the feature patch map and shrinking the responses at invisible points, leading to the Pose Adaptive Feature (PAF).}
  \label{fig-PAF}
\end{figure}

Traditional convolutional layers convolve along a 2D map from pixel to pixel, while we intend to convolve at some semantically consistent locations on the face, called Pose Adaptive Convolution (PAC). Considering human face can be roughly approximated with a cylinder~\cite{Spreeuwers-2011-cylinder}, we compute the cylindrical coordinate of each vertex and sample $64 \times 64$ feature anchors with constant azimuth and height intervals, see Fig.~\ref{fig-PAF-a}.

Given a current model parameter $\mathbf{p}$, we first project 3DMM and sample the feature anchors on the image plane, getting $64 \times 64 \times 2$ projected feature anchors $V(\mathbf{p})_{anchor}$ (Fig.~\ref{fig-PAF-b}). Second we crop $d \times d$ ($5$ in our implementation) patch at each feature anchor and concatenate the patches into a $(64*d) \times (64*d)$ patch map according to their cylindrical coordinates (Fig.~\ref{fig-PAF-c}). Finally we conduct $d \times d$ convolutions at the stride of $d$ on the patch map, generating $64 \times 64$ response maps (Fig.~\ref{fig-PAF-d}). The convolutional filters are learned with a common convolutional layer, jointly with other CNN layers as described in Sec.~\ref{sec-network-structure}.

Note that this process is equivalent to directly conducting $d \times d$ convolutions on the projected feature anchors $V(\mathbf{p})_{anchor}$, which implicitly localize and frontalize the face, making the convolution pose invariant. In order to shrink the features at the occluded region, we consider the vertices whose normal points to minus $z$ as self-occluded and divide the responses at occluded region by two, generating the Pose Adaptive Feature (PAF). We do not eliminate occluded features as~\cite{jourabloo2015pose} since this information is still valuable prior to perfect fitting.

\subsubsection{Projected Normalized Coordinate Code}
\begin{figure}[!htb]
  \centering
  \subfigure[NCC]{\label{fig-PNCC-a}
  \includegraphics[width=0.225\textwidth]{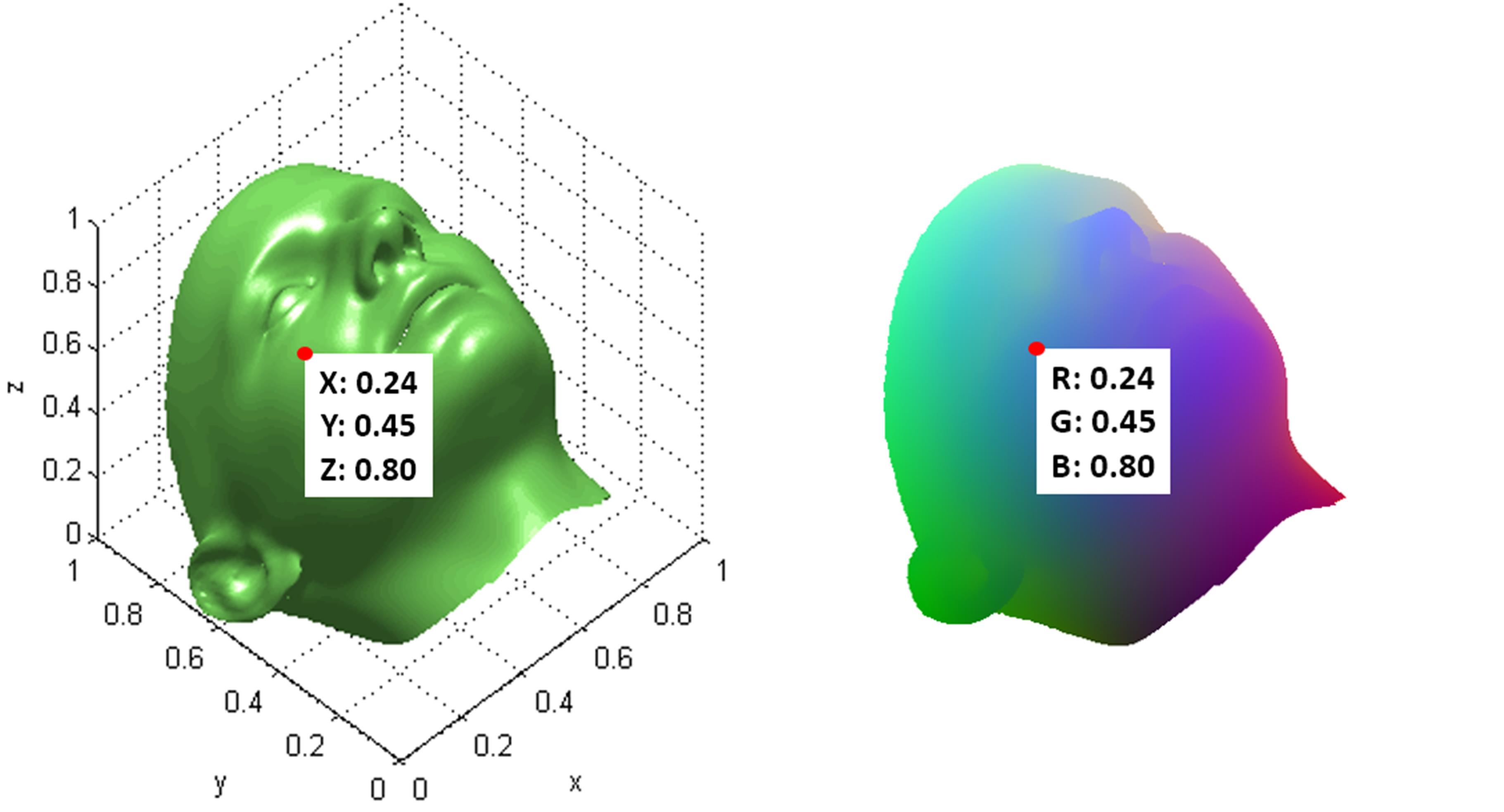}}
  \subfigure[PNCC]{\label{fig-PNCC-b}
  \includegraphics[width=0.225\textwidth]{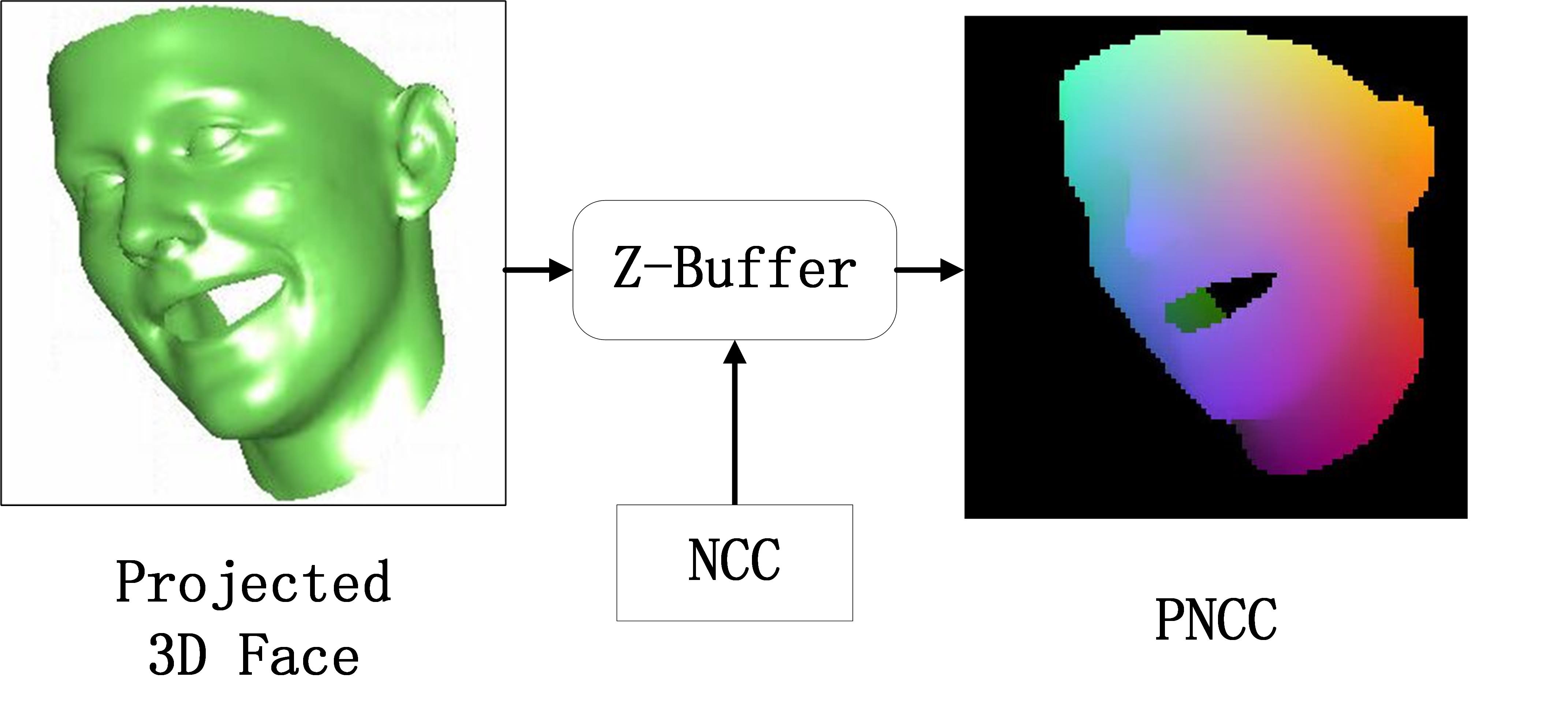}}
  \caption{The Normalized Coordinate Code (NCC) and the Projected Normalized Coordinate Code (PNCC). (a) The normalized mean face, which is also demonstrated with NCC as its texture ($\text{NCC}_{x}=\text{R}$, $\text{NCC}_{y}=\text{G}$, $\text{NCC}_{z}=\text{B}$). (b) The generation of PNCC, the projected 3D face is rendered by Z-Buffer with NCC as its colormap.}
  \label{fig-PNCC}
\end{figure}

The proposed image-view feature depends on a new type of vertex index, which is introduced as follows: we normalize the 3D mean face to $0-1$ in $x,y,z$ axis as Eqn.~\ref{equ-NCC}:
\begin{equation}\label{equ-NCC}
  \text{NCC}_{d}=\frac{\mathbf{\overline{S}}_{d} - \min(\overline{\mathbf{S}}_{d})}{\max(\overline{\mathbf{S}}_{d}) - \min(\overline{\mathbf{S}}_{d})}~~~ (d = x,y,z),
\end{equation}
where the $\mathbf{\overline{S}}$ is the mean shape of 3DMM. After normalization, the 3D coordinate of each vertex \textbf{uniquely} distributes between $[0,0,0]$ and $[1,1,1]$, so it can be considered as a vertex index, which we call Normalized Coordinate Code (NCC) (Fig.~\ref{fig-PNCC-a}). Since NCC has three channels as RGB, we can also show NCC as the face texture. It can be seen as different from the traditional vertex index (from $1$ to the number of vertices), NCC is smooth along the face surface.

In the fitting process, with a model parameter $\mathbf{p}$, we adopt Z-Buffer to render the projected 3D face colored by NCC (Fig.~\ref{fig-PNCC-b}) as in Eqn.~\ref{equ-PNCC}:
\begin{gather}
   \text{PNCC}= \emph{Z-Buffer}(V_{3d}(\mathbf{p}),  \text{NCC}), \label{equ-PNCC}\\
  V_{3d}(\mathbf{p}) = \mathbf{R} * (\mathbf{\overline{S}} + \mathbf{A}_{id}\bm{\alpha}_{id} + \mathbf{A}_{exp}\bm{\alpha}_{exp}) + [\mathbf{t}_{2d},0]^\mathrm{T}, \notag
\end{gather}
where $\emph{Z-Buffer}(\bm{\nu},\bm{\tau})$ renders the 3D mesh $\bm{\nu}$ colored by $\bm{\tau}$ and $V_{3d}(\mathbf{p})$ is the projected 3D face. We call the rendered image Projected Normalized Coordinate Code (PNCC). Afterwards, PNCC is stacked with the input image and sent to the CNN.

Comparing PAF and PNCC, we can see that PAF is a model-view feature since it implicitly warps the image with feature anchors and PNCC is an image-view feature it sends the original image into a CNN. Regarding the three properties, they fulfill the feedback property since they both depend on $\mathbf{p}$ which is updated by the output of the CNN. As for the convolvable property, PAC is the convolution on the continuous locations indicated by the feature anchors and its result PAF is a smooth 2D map. PNCC is also smooth in 2D and the convolution indicates the linear combination of NCCs on a local patch. As for the convergence property, when the CNN detects that in PAF the face is aligned to front and in PNCC each NCC superposes its corresponding image pattern, the cascade will converge.

\subsection{Network Structure}\label{sec-network-structure}
Unlike existing CNN methods~\cite{sun2013deep,liang2015unconstrained} that apply different network structures for different fitting stages, 3DDFA employs a unified network structure across the cascade. In general, at iteration $k$ ($k=0,1,...,K$), given an initial parameter $\mathbf{p}^{k}$, we construct PNCC and PAF with $\mathbf{p}^{k}$ and train a two-stream CNN $\emph{Net}^{k}$ to conduct fitting. The output features from two streams are merged to predict the parameter update $\Delta \mathbf{p}^{k}$:
\begin{equation}\label{equ-3DDFA}
  \Delta \mathbf{p}^{k} = \emph{Net}^{k} (\text{PAF}(\mathbf{p}^{k}, \mathbf{I}), \text{PNCC}(\mathbf{p}^{k}, \mathbf{I})).
\end{equation}
Afterwards, a better intermediate parameter $\mathbf{p}^{k+1}=\mathbf{p}^{k}+\Delta \mathbf{p}^{k}$ becomes the input of the next network $\emph{Net}^{k+1}$ which has the same structure but different weights with $\emph{Net}^{k}$.
Fig.~\ref{fig-overview} shows the network structure. In the PNCC stream, the input is the $200\times200\times3$ color image stacked by the $200\times200\times3$ PNCC. The network contains five convolutional layers, four pooling layers and one fully connected layer. In the PAF stream, the input is the $200\times200\times3$ color image and $64\times64$ feature anchors. The image is processed with the pose adaptive convolution, followed by three pooling layers, three convolutional layers and one fully connected layer. The outputs of the two streams are merged with an additional fully connected layer to predict the $234$-dimensional parameter update including $6$-dimensional pose parameters $[q_{0}, q_{1}, q_{2}, q_{3}, t_{2dx}, t_{2dy}]$, $199$-dimensional shape parameters $\bm{\alpha}_{id}$ and $29$-dimensional expression parameters $\bm{\alpha}_{exp}$.

\subsection{Cost Function}\label{sec-cost-function}
Different from the landmark shape model, the parameters in 3DMM contribute to the fitting accuracy with very different impacts, giving parameters different priorities. As a result, regression-based methods suffer from the inequivalence between parameter error and alignment error~\cite{Cao-CVPR-12}. In this section, we will discuss this problem with two baseline cost functions and propose our own ways to model the parameter priority.

\subsubsection{Parameter Distance Cost (PDC)}
Take the first iteration as an example. The purpose of the CNN is to predict the parameter update $\Delta \mathbf{p}$ so as to move the initial parameter $\mathbf{p}^{0}$ closer to the ground truth $\mathbf{p}^{g}$.
Intuitively, we can minimize the distance between the ground truth and the current parameter with the Parameter Distance Cost (PDC):
\begin{equation}\label{equ-pdc}
  E_{pdc} = \| \Delta \mathbf{p} - (\mathbf{p}^{g}-\mathbf{p}^{0}) \|^{2}.
\end{equation}
PDC has been traditionally used in regression based model fitting~\cite{hou2001direct, saragih2007nonlinear, zhu2015discriminative}. However, different dimension in $\mathbf{p}$ has different influences on the resultant 3D face. For example, with the same deviation, the yaw angle will bring a larger alignment error than a shape parameter, while PDC optimizes them equally, leading to sub-optimal results.

\subsubsection{Vertex Distance Cost (VDC)}
Since 3DDFA aims to morph the 3DMM to the ground truth 3D face, we can optimize $\Delta \mathbf{p}$ by minimizing the vertex distances between the current and the ground truth 3D face:
\begin{equation}\label{equ-vdc}
  E_{vdc} = \| V(\mathbf{p}^{0} + \Delta \mathbf{p}) - V(\mathbf{p}^{g}) \|^{2},
\end{equation}
where $V(\cdot)$ is the face construction and projection as Eqn.~\ref{equ-projection}. We call this cost Vertex Distance Cost (VDC).
Compared with PDC, VDC better models the fitting error by explicitly considering parameter semantics. However, VDC is not convex itself, the optimization is not guaranteed to converge to the ground truth parameter $\mathbf{p}^{g}$. Furthermore, we observe that VDC exhibits pathological curvature~\cite{martens2010deep} since the directions of pose parameters always exhibit much higher curvatures than the PCA coefficients. As a result, optimizing VDC with gradient descent converges very slowly due to the ``zig-zagging'' problem. Second-order optimizations are preferred to handle the pathological curvature but they are expensive and hard to be implemented on GPU.

\subsubsection{Weighted Parameter Distance Cost (WPDC)}
In our previous work~\cite{zhu2016face}, we propose a cost function named Weighted Parameter Distance Cost (WPDC). The motivation is explicitly weighting parameter error by its importance:
\begin{equation}
  E_{wpdc} = (\Delta \mathbf{p} - (\mathbf{p}^{g}-\mathbf{p}^{0}))^\mathrm{T}\mathrm{diag}(\mathbf{w})(\Delta \mathbf{p} - (\mathbf{p}^{g}-\mathbf{p}^{0}))
\label{equ-wpdc}
\end{equation}
where $\mathbf{w}$ is the parameter importance vector, which is defined as follows:
\begin{equation}
\begin{split}
  \mathbf{w} = (w_{1},w_{2}&,...,w_{i},...,w_{p}),\\
  w_{i} = \| V(\mathbf{p}^{de,i}&) - V(\mathbf{p}^{g}) \| / Z,\\
  \mathbf{p}^{de,i} = (\mathbf{p}^{g}_{1},...,\mathbf{p}^{g}_{i-1},&(\mathbf{p}^{0}+\Delta \mathbf{p})_{i},\mathbf{p}^{g}_{i+1},...,\mathbf{p}^{g}_{p}),\\
\label{equ-wpdc-importance}
\end{split}
\end{equation}
where $p$ is the number of parameter, $\mathbf{p}^{de,i}$ is the $i$-degraded parameter whose $i$th element comes from the predicted parameter $(\mathbf{p}^{0}+\Delta \mathbf{p})$ and the others come from the ground truth parameter $\mathbf{p}^{g}$, $Z$ is a regular term which is the maximum of $\mathbf{w}$. $\| V(\mathbf{p}^{de}(i)) - V(\mathbf{p}^{g}) \|$ models the alignment error brought by miss-predicting the $i$th model parameter, which is indicative of its importance. In the training process, the CNN firstly concentrates on the parameters with larger $\| V(\mathbf{p}^{de}(i)) - V(\mathbf{p}^{g}) \|$ such as rotation and translation. As $\mathbf{p}^{de}(i)$ is closer to $\mathbf{p}^{g}$, the weights of these parameters begin to shrink and the CNN will optimize less important parameters while simultaneously keeping the high-priority parameters sufficiently good. Compared with VDC, WPDC makes sure the parameter is optimized toward $\mathbf{p}^{g}$ and it remedies the pathological curvature issue at the same time.

\begin{figure}[!htb]
  \centering
  \subfigure[Image]{\label{fig-WPDC-fail-a}
  \includegraphics[width=0.12\textwidth]{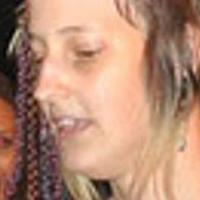}}
  \subfigure[Error = 10.72]{\label{fig-WPDC-fail-b}
  \includegraphics[width=0.12\textwidth]{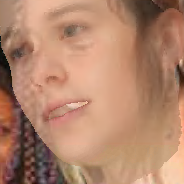}}
  \subfigure[Error = 4.39]{\label{fig-WPDC-fail-c}
  \includegraphics[width=0.12\textwidth]{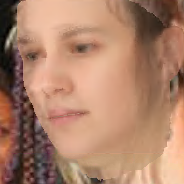}}
  \caption{(a) An open-mouth face in near-profile view. (b) The fitting result of WPDC in the first iteration. (c) The fitting result when the CNN is restricted to only regress the $6$-dimensional pose parameters. Errors are measured by Normalized Mean Error.}
  \label{fig-WPDC-fail}
\end{figure}

However, the weight in WPDC only models the ``importance'' but not the ``priority''. In fact, parameters become important sequentially. Take Fig.~\ref{fig-WPDC-fail} as an example, when WPDC evaluates a face image with open mouth and large pose, it will assign both expression and rotation high weights. We can observe that attempting to estimate expression makes little sense before the pose is accurate enough, see Fig.~\ref{fig-WPDC-fail-b}. One step further, if we force the CNN to only concentrate on pose parameters, we obtain a better fitting result, see Fig.~\ref{fig-WPDC-fail-c}. Consequently for this sample, even though pose and expression are both important, pose has higher priority than expression, but WPDC misses that.

\subsubsection{Optimized Weighted Parameter Distance Cost (OWPDC)}
We can observe that ``priority'' is a between-parameter relationship which can only be modeled by treating all the parameters as a whole rather than evaluating them separately as WPDC. In this paper, we propose to find the best weights through optimization:
\begin{align}\label{equ-owpdc-cost}
E_{owpdc} = (\Delta \mathbf{p} - (\mathbf{p}^{g}-\mathbf{p}^{0}))&^\mathrm{T}\mathrm{diag}(\mathbf{w}^{*})(\Delta \mathbf{p} - (\mathbf{p}^{g}-\mathbf{p}^{0})), \notag\\
  \mathbf{w}^{*} = \arg \min\limits_{\mathbf{w}} \Big\|V \Big(\mathbf{p}^{c} + \mathrm{diag}&(\mathbf{w})*(\mathbf{p}^{g} - \mathbf{p}^{c})\Big) - V\Big(\mathbf{p}^{g}\Big) \Big\|^{2} \notag \\
  + \lambda~\Big\|\mathrm{diag}(\mathbf{w})*&(\mathbf{p}^{g} - \mathbf{p}^{c})\Big\|^{2}, \\
  s.t. ~~~~~ \mathbf{0} \preceq ~&\mathbf{w} \preceq ~\mathbf{1}, \notag
\end{align}
where $\mathbf{w}$ is the weights vector, $\Delta \mathbf{p}$ is the CNN output, $\mathbf{p}^{c}=\mathbf{p}^{0} + \Delta \mathbf{p}$ is the current predicted parameter, $\mathbf{0}$ and $\mathbf{1}$ are the zeros and ones vectors respectively and $\preceq$ is the element-wise less than. In Eqn.~\ref{equ-owpdc-cost}, by adding a weighted parameter update $\mathrm{diag}(\mathbf{w})(\mathbf{p}^{g} - \mathbf{p}^{c})$ to the current parameter $\mathbf{p}^{c}$, we hope the new face is closer to the ground truth face with limited updating.
Note that $\|\mathrm{diag}(\mathbf{w})*(\mathbf{p}^{g} - \mathbf{p}^{c})\|^{2}$ is the square sum of the gradient of OWPDC, which models how much CNN weights need to be tuned to predict each parameter. We use this penalty term to choose the parameters which are most beneficial to the fitting and are easiest to learn. The range of $\mathbf{w}$ is constrained to be $[0,1]$ to make sure the parameter is optimized to $\mathbf{p}^{g}$. Obviously, when the $\lambda$ is set to 0, there will be a trivial solution that $\mathbf{w}=\mathbf{1}$ and OWPDC will deteriorate to PDC.

In the training process, directly optimizing Eqn.~\ref{equ-owpdc-cost} for each sample is computationally intensive. We expand $V (\mathbf{p}^{c} + \mathrm{diag}(\mathbf{w})(\mathbf{p}^{g} - \mathbf{p}^{c}))$ at $\mathbf{p}^{g}$ with the Taylor formula and let $\Delta \mathbf{p}^{c}= \mathbf{p}^{g} - \mathbf{p}^{c}$, Eqn.~\ref{equ-owpdc-cost} will be:
\begin{equation}
\centering
\Big\|V'(\mathbf{p}^{g})*\mathrm{diag}(\mathbf{w}-\mathbf{1})*\Delta \mathbf{p}^{c}\Big\|^{2} + \lambda~\Big\|\mathrm{diag}(\mathbf{w})*\Delta \mathbf{p}^{c}\Big\|^{2},
\label{equ-owpdc-talor}
\end{equation}
where $V'(\mathbf{p}^{g})$ is the Jacobian. Expanding Eqn.~\ref{equ-owpdc-talor} and removing the constant terms, we get:
\begin{align}\label{equ-owpdc-talor-expand}
\mathbf{w}^\mathrm{T} \Big( \mathrm{diag}(\Delta \mathbf{p}^{c})V'(\mathbf{p}^{g})&^\mathrm{T}V'(\mathbf{p}^{g})\mathrm{diag}(\Delta \mathbf{p}^{c}) \Big)\mathbf{w} \notag\\
- 2 * \mathbf{1}^\mathrm{T} \Big( \mathrm{diag}(\Delta \mathbf{p}^{c})V'(\mathbf{p}^{g})&^\mathrm{T}V'(\mathbf{p}^{g})\mathrm{diag}(\Delta \mathbf{p}^{c}) \Big)\mathbf{w} \notag\\
+ \mathbf{\lambda}* \mathbf{w}^\mathrm{T}\mathrm{diag}(&\Delta \mathbf{p}^{c}.*\Delta \mathbf{p}^{c})\mathbf{w},
\end{align}
where $.*$ is the element-wise multiplication. Let $\mathbf{H}=V'(\mathbf{p}^{g})\mathrm{diag}(\Delta \mathbf{p}^{c})$ which is a $2n \times p$ matrix where $n$ is the number of vertices and $p$ is the number of parameters, the optimization will be:
\begin{align}\label{equ-owpdc-final}
\arg \min\limits_{\mathbf{w}}\mathbf{w}^\mathrm{T}*(\mathbf{H}^\mathrm{T}*\mathbf{H}+\lambda*&\mathrm{diag}(\Delta \mathbf{p}^{c}.*\Delta \mathbf{p}^{c}))*\mathbf{w} \notag\\
+2*\mathbf{1}^\mathrm{T}*\mathbf{H}&^\mathrm{T}*\mathbf{H}*\mathbf{w}, \notag\\
s.t. ~~~~~ \mathbf{0} \preceq ~&\mathbf{w} \preceq ~\mathbf{1},
\end{align}
which is a standard quadratic programming problem with the unique solution. The most consuming component in Eqn.~\ref{equ-owpdc-final} is the computation of $V'(\mathbf{p}^{g})$. Fortunately, $\mathbf{p}^{g}$ is constant during training and $V'(\mathbf{p}^{g})$ can be pre-computed offline. As a result, the computation of $\mathbf{w}^{*}$ can be reduced to a $p$-dimensional quadratic programming which can be efficiently solved.
The only parameter in OWPDC is the $\lambda$. It directly determines which parameter is valid during training. We set $\lambda=0.17*\|V (\mathbf{p}^{c}) - V(\mathbf{p}^{g}) \|^{2}$ in our implementation.

\section{Face Profiling}
All the regression based methods rely on training data, especially for CNNs which have thousands of parameters to learn. Therefore, massive labelled faces in large poses are crucial for 3DDFA. However, few of the released face alignment databases contain large-pose samples~\cite{zhu2012face,jaiswal2013guided,le2012interactive,sagonas2013semi} since labelling standardized landmarks on them is very challenging. In this work, we demonstrate that profile faces can be well synthesized from existing training samples with the help of 3D information. Inspired by the recent achievements in face frontalization~\cite{zhu2015high,hassner2014effective} which generates the frontal view of faces, we propose to invert this process to synthesize the profile view of faces from medium-pose samples, which is called face profiling. Different from the face synthesizing in recognition~\cite{Prabhu-PAMI-2011}, face profiling is not required to keep the identity information but to make the synthesizing results realistic. However, current synthesizing methods do not keep the external face region~\cite{masi2016pose,Prabhu-PAMI-2011}, which contains important context information for face alignment. In this section, we elucidate a novel face synthesizing method to generate the profile views of face image with out-of-plane rotation, providing abundant realistic training samples for 3DDFA.

\subsection{3D Image Meshing}
\begin{figure}[!htb]
  \centering
  \subfigure[]{
  \label{fig-3D-meshing-a}
  \includegraphics[width=0.11\textwidth]{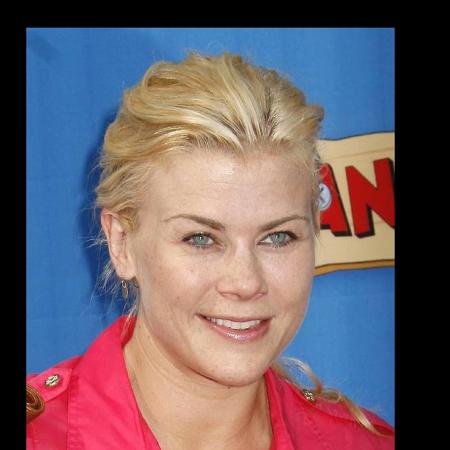}}
  \subfigure[]{
  \label{fig-3D-meshing-b}
  \includegraphics[width=0.11\textwidth]{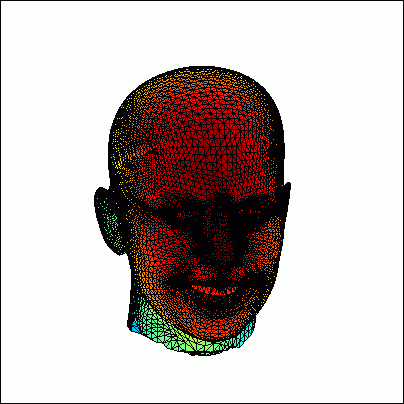}}
  \subfigure[]{
  \label{fig-3D-meshing-c}
  \includegraphics[width=0.11\textwidth]{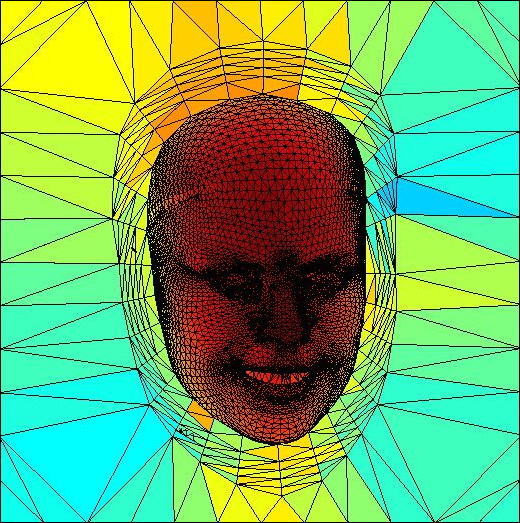}}
  \subfigure[]{
  \label{fig-3D-meshing-d}
  \includegraphics[width=0.11\textwidth]{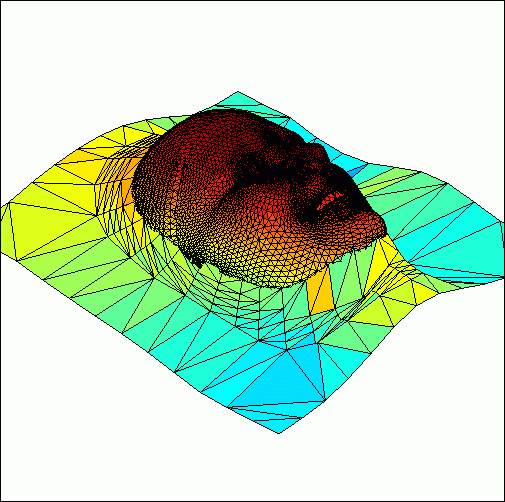}}
  \caption{3D Image Meshing. (a) The input image. (b) The fitted 3D face through MFF. (c) The depth image from 3D meshing. (d) A different view of the depth image.}
  \label{fig-3D-meshing}
\end{figure}

The depth estimation of a face image can be conducted on the face region and the external region respectively, with different requirements of accuracy. On the face region, we fit a 3DMM through the Multi-Features Framework (MFF)~\cite{Romdhani-CVPR-05} (see Fig.~\ref{fig-3D-meshing-b}). With the ground truth landmarks as a solid constraint throughout the fitting process, MFF can always get accurate results. Few difficult samples can be easily adjusted manually. On the external region, we follow the 3D meshing method proposed by Zhu et al.~\cite{zhu2015high} to mark some anchors beyond the face region and simulate their depth, see Fig.~\ref{fig-3D-meshing-c}. Afterwards the whole image can be tuned into a 3D object through triangulation (see Fig.~\ref{fig-3D-meshing-c}\ref{fig-3D-meshing-d}).

\subsection{3D Image Rotation}
\begin{figure}[!htb]
  \centering
  \subfigure{
  \label{fig-anchor-adjust-a}}
  \subfigure{
  \label{fig-anchor-adjust-b}}
  \subfigure{
  \label{fig-anchor-adjust-c}}
  \includegraphics[width=0.45\textwidth]{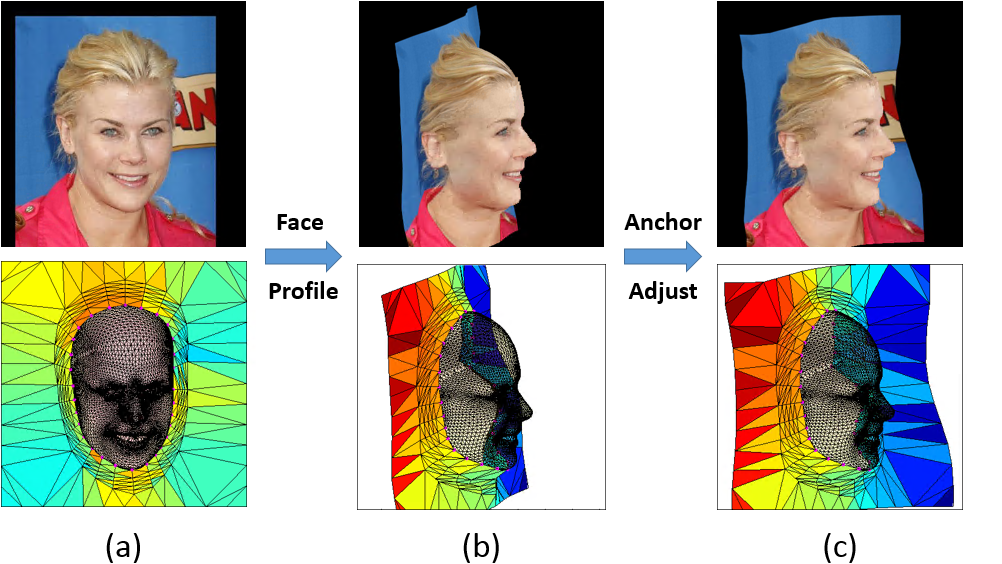}
  \caption{The face profiling and anchor adjustment process. (a) The source image. (b) The profiled face with out of plane rotation. It can be seen that the face locates on the hollow since the background is squeezed. (c) The synthesized image after anchor adjustment.}
  \label{fig-anchor-adjust}
\end{figure}

The simulated depth information enables the 2D image to rotate out of plane to generate the appearances in larger poses. However, as shown in Fig.~\ref{fig-anchor-adjust-b}, the 3D rotation squeezes the external face region and loses the background. As a result, we need to further adjust the anchors to keep the background relatively unchanged and preserve the smoothness simultaneously. Inspired by our previous work~\cite{zhu2015high}, we propose to adjust background anchors by solving an equation list about their relative positions.

In the source image as shown in Fig.~\ref{fig-anchor-adjust-a}, the triangulated anchors build up a graph where the anchors are the vertices and the mesh lines are the edges. In the graph, each edge represents an anchor-to-anchor relationship:
\begin{equation}\label{equ-anchor-adjust1}
  x_{a\_src}-x_{b\_src}=\Delta x_{src}, ~~~~~~~
  y_{a\_src}-y_{b\_src}=\Delta y_{src},
\end{equation}
where $(x_{a\_src},y_{a\_src})$ and $(x_{b\_src},y_{b\_src})$ are two connecting anchors, $\Delta x_{src}$ and $\Delta y_{src}$ are the spatial offsets in $x$, $y$ axes, which should be preserved in synthesizing. After profiling, we keep the face contour anchors (the magenta points in Fig.~\ref{fig-anchor-adjust-b}) consistent and predicting other anchors with the unchanged anchor offsets:
\begin{equation}\label{equ-anchor-adjust2}
  x_{a\_adj}-x_{b\_adj}= \Delta x_{src}, ~~~~
  y_{a\_adj}-y_{b\_adj}= \Delta y_{src},
\end{equation}
Specifically, if $a$ is a face contour anchor, we set $(x_{a\_adj},y_{a\_adj})$ to the positions after profiling $(x_{a\_pro},y_{a\_pro})$, otherwise $(x_{a\_adj},y_{a\_adj})$ are two unknowns need to be solved. By collecting Eqn.~\ref{equ-anchor-adjust2} for each graph edge, we form an equation list whose least square solution is the adjusted anchors (as seen in Fig.~\ref{fig-anchor-adjust-c}).

\begin{figure}[!htb]
  \centering
  \subfigure{
  \includegraphics[width=0.11\textwidth]{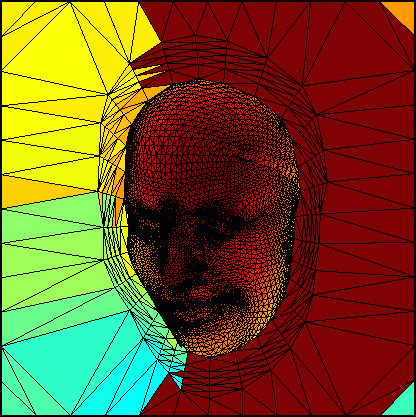}}
  \subfigure{
  \includegraphics[width=0.11\textwidth]{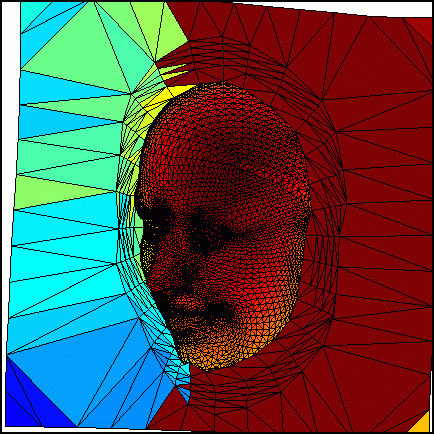}}
  \subfigure{
  \includegraphics[width=0.11\textwidth]{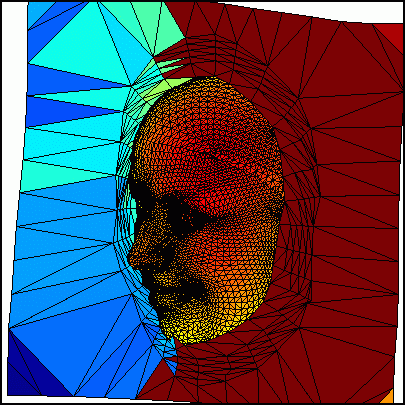}}
  \subfigure{
  \includegraphics[width=0.11\textwidth]{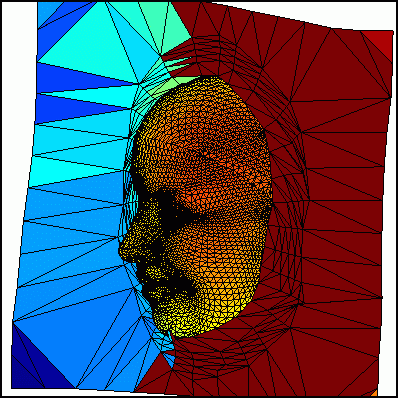}}\\
  \setcounter{subfigure}{0}
  \subfigure[]{
  \label{fig-img-meshing-a}
  \includegraphics[width=0.11\textwidth]{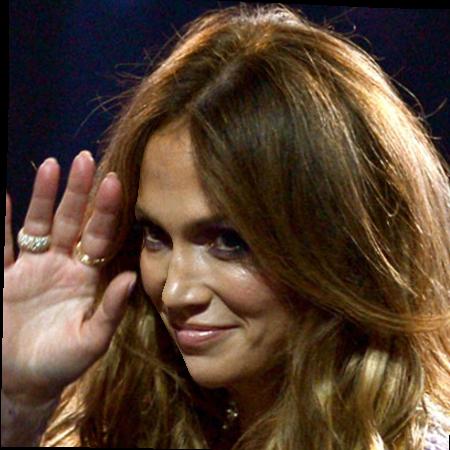}}
  \subfigure[]{
  \label{fig-img-meshing-b}
  \includegraphics[width=0.11\textwidth]{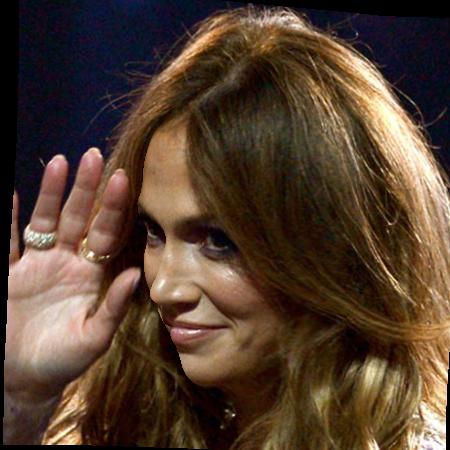}}
  \subfigure[]{
  \label{fig-img-meshing-c}
  \includegraphics[width=0.11\textwidth]{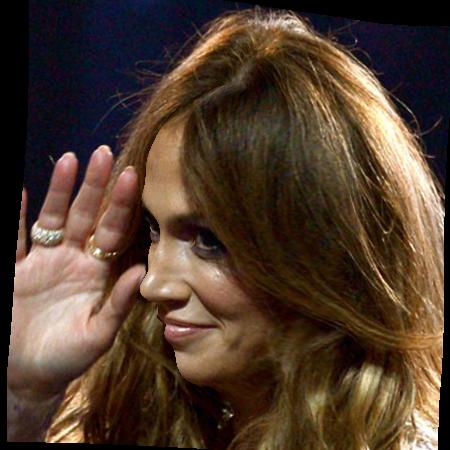}}
  \subfigure[]{
  \label{fig-img-meshing-d}
  \includegraphics[width=0.11\textwidth]{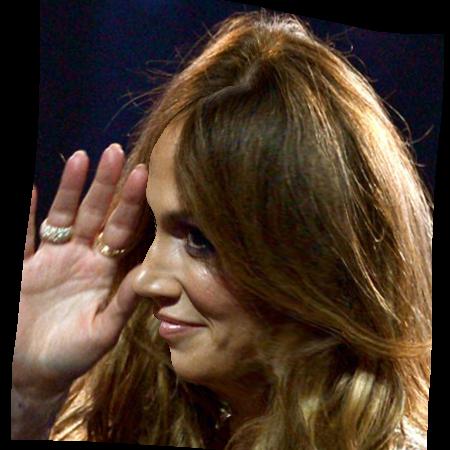}}
  \caption{2D and 3D view of face profiling. (a) The original yaw angle $yaw_{0}$. (b) $ yaw_{0}+ 20^{\circ}$. (c) $yaw_{0} + 30^{\circ}$. (d) $yaw_{0} + 40^{\circ}$.}
  \label{fig-img-rotation}
\end{figure}

In this work, we enlarge the $yaw$ angle of image at the step of $5^{\circ}$ until $90^{\circ}$, see Fig.~\ref{fig-img-rotation}. Different from face frontalization, with larger rotation angles the self-occluded region can only be expanded. As a result, we avoid the troubling invisible region filling which may produce large artifacts~\cite{zhu2015high}. Through face profiling, we not only obtain face samples in large poses but also augment the dataset to a large scale.

\section{Implementation}
\vspace{0.3em}
\noindent\textbf{Training Strategy}:
With a huge number of parameters, the CNN tends to overfit the training set and the deeper cascade might learn nothing with overfitted samples. Therefore we regenerate $\mathbf{p}^{k}$ at each iteration using a nearest neighbor strategy. By observing that the fitting error highly depends on the ground truth face posture (FP), we perturb a training sample based on a set of similar-FP validation samples. In this paper, we define the face posture as the rotated 3D face without scaling and translation:
\begin{equation}\label{equ-model-condition}
  \text{FP}=\mathbf{R}^{g}*(\mathbf{\overline{S}} + \mathbf{A}_{id}\bm{\alpha}_{id}^{g} + \mathbf{A}_{exp}\bm{\alpha}_{exp}^{g}),
\end{equation}
where $\mathbf{R}^{g}$ is constructed from the normalized ground truth quaternion, $\bm{\alpha}_{id}^{g}$ and $\bm{\alpha}_{exp}^{g}$ are the ground truth shape and expression parameters respectively.
Before training, we select two folds of samples as the validation set and for each training sample we construct a validation subset $\{v_{1},...,v_{m}\}$ whose members share similar FP with the training sample. At iteration $k$, we regenerate the initial parameter by:
\begin{equation}\label{equ-init-regeneration}
  \mathbf{p}^{k} = \mathbf{p}^{g} - (\mathbf{p}^{g}_{v_{i}} - \mathbf{p}^{k}_{v_{i}}),
\end{equation}
where $\mathbf{p}^{k}$ and $\mathbf{p}^{g}$ are the initial and ground truth parameter of a training sample,  $\mathbf{p}^{k}_{v_{i}}$ and $\mathbf{p}^{g}_{v_{i}}$ come from a validation sample $v_{i}$ which is randomly chosen from the corresponding validation subset. Note that $v_{i}$ is never used in training.

\vspace{0.3em}
\noindent\textbf{Initialization}:
Besides the face profiling, we also augment the training data ($10$ times) by randomly in-plane rotating images (up to $30$ degrees) and perturbing bounding boxes. Specifically, the bounding boxes are randomly perturbed by a multivariate normal distribution whose mean vector and covariance matrix are obtained by the difference between ground truth bounding boxes and automated detected face rectangles using FTF~\cite{peiyun2017finding}. This augmentation is quite effective in improving the robustness of the model. During testing, to get $\mathbf{p}^{0}$ we first set $\bm{\alpha}_{id}$, $\bm{\alpha}_{exp}$ to zero and the quaternion to $[1,0,0,0]$, getting a frontal 3D mean face. Then we calculate $\mathbf{t}_{2d}$ by moving the mean point of the 3D face to the center of the bounding box. Finally, we scale the 3D face, which is equivalent to scaling the quaternion, to make the bounding box enclose the whole face region.

\vspace{0.3em}
\noindent\textbf{Running Time}:
During testing, 3DDFA takes $21.3$ms for each iteration, among which PAF and PNCC take $11.6$ms and $6.8$ms respectively on $3.40$GHZ CPU and CNN forward propagation takes $2.9$ms on GTX TITAN X GPU. In our implementation, 3DDFA has three iterations and takes $63.9$ms ($15.65$fps) for each sample. Note that the efficiency is mainly limited by the input features, which can be further improved by GPU implementation.

\section{Experiments}
\subsection{Datasets}\label{sec-datasets}
Three databases are used in our experiments, i.e. 300W-LP, AFLW~\cite{kostinger2011annotated} and a specifically constructed AFLW2000-3D.

\textbf{300W-LP}: 300W~\cite{sagonas2013300} standardises multiple face alignment databases with 68 landmarks, including AFW~\cite{zhu2012face}, LFPW~\cite{Belhumeur-2011-LFPW}, HELEN~\cite{zhou2013extensive}, IBUG~\cite{sagonas2013300} and XM2VTS~\cite{messer1999xm2vtsdb}. With 300W, we adopt the proposed face profiling to generate $61,225$ samples across large poses ($1,786$ from IBUG, $5,207$ from AFW, $16,556$ from LFPW and $37,676$ from HELEN, XM2VTS is not used), which is further flipped to $122,450$ samples. We call the synthesized database as 300W Across Large Poses (300W-LP).

\textbf{AFLW}: AFLW~\cite{kostinger2011annotated} contains $21,080$ in-the-wild faces with large pose variations (yaw from $-90^{\circ}$ to $90^{\circ}$). Each image is annotated up to 21 visible landmarks. The database is very suitable for evaluating face alignment performance in large poses.

\begin{figure}[!htb]
  \centering
  \includegraphics[width=0.45\textwidth]{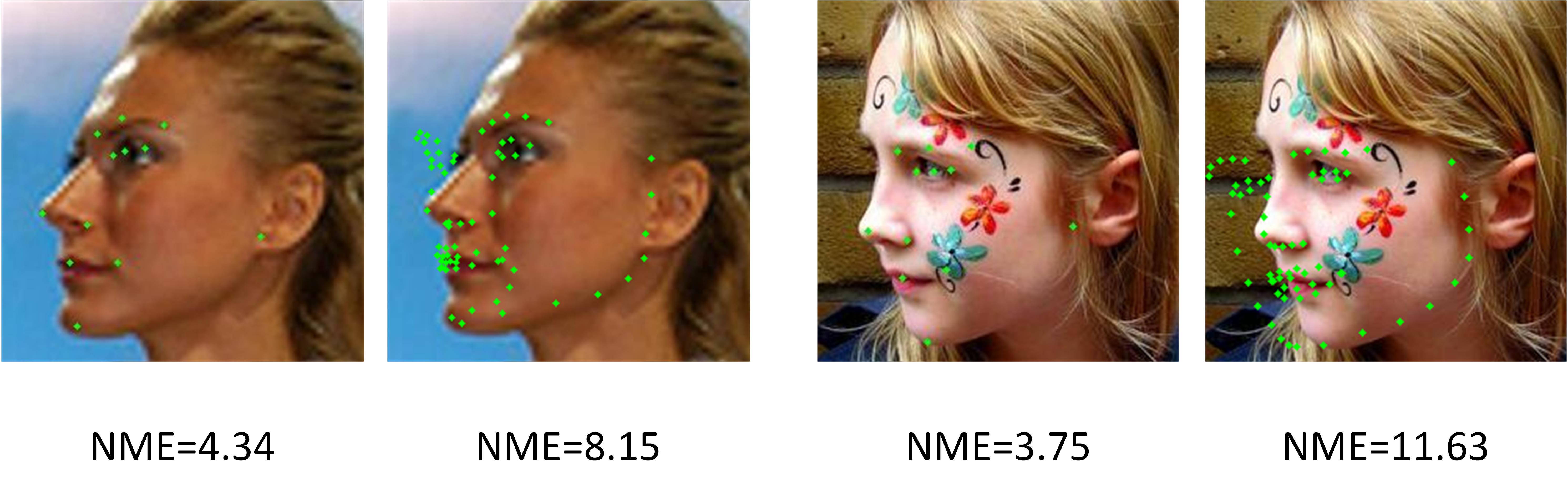}
  \caption{Fake good alignment in AFLW. For each sample, the first shows the visible 21 landmarks and the second shows all the $68$ landmarks. The Normalized Mean Error (NME) reflects their accuracy. It can be seen that only evaluating visible landmarks cannot well reflect the accuracy of 3D fitting.}
  \label{fig-fakegood}
\end{figure}

\textbf{AFLW2000-3D}: Evaluating 3D face alignment in the wild is difficult due to the lack of pairs of 2D image and 3D scan. Considering the recent achievements in 3D face reconstruction which can construct a 3D face from 2D landmarks~\cite{Aldrian-PAMI-13,zhu2015high}, we assume that a 3D model can be accurately fitted if sufficient 2D landmarks are provided. Therefore the evaluation can be degraded to 2D landmark evaluation which also makes it possible to compare 3DDFA with other 2D face alignment methods. While AFLW is not suitable for this task since only visible landmarks may lead to serious ambiguity in 3D shape, as reflected by the fake good alignment phenomenon in Fig.~\ref{fig-fakegood}. In this work, we construct a database called AFLW2000-3D for 3D face alignment evaluation, which contains the ground truth 3D faces and the corresponding 68 landmarks of the first 2,000 AFLW samples. More details about the construction of AFLW2000-3D are given in supplemental material.

In all the following experiments, we follow \cite{zhu2015face} and regard the 300W-LP samples synthesized from the training part of LFPW, HELEN and the whole AFW as the training set ($101,144$ images in total). The testing are conducted on three databases: the 300W testing part for general face alignment, the AFLW for large-pose face alignment and the AFLW2000-3D for 3D face alignment. The alignment accuracy is evaluated by the Normalized Mean Error (NME).

\subsection{Performance with Different Input Features}
As described in Sec.~\ref{sec-feature-design}, the input features of face alignment methods can be divided into two categories, the image-view feature and the model-view feature, which correspond to PNCC and PAF in this paper. To test their effectiveness respectively and evaluate their complementarity, we divide the network in Fig.~\ref{fig-overview} into PNCC stream and PAF stream by removing the last fully connected layer and regress the $256$-dimensional output of each stream to the parameter update respectively. The combined two-stream network is also reported to demonstrate the improvements.
\begin{figure}[!htb]
  \centering
  \includegraphics[width=0.4\textwidth]{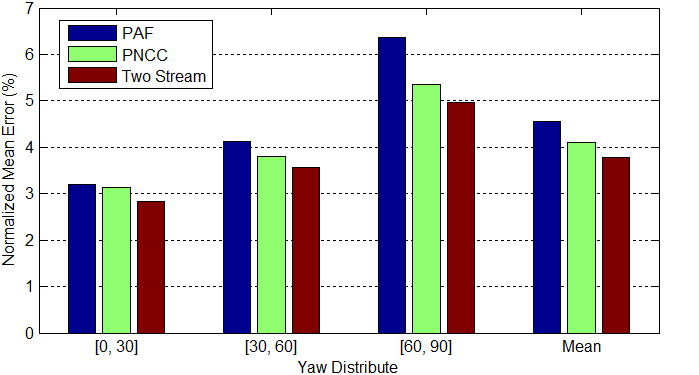}
  \caption{The Normalized Mean Error ($\%$) with different input features, evaluated on AFLW2000-3D with different yaw intervals.}
  \label{fig-different-feature}
\end{figure}
As shown in Fig.~\ref{fig-different-feature}, PNCC performs better than PAF when used individually and the improvement is enlarged as the pose becomes larger. Besides, PNCC and PAF achieve better performance when combined, which may infer a complementary relationship. This complementary relationship might be because PNCC covers the whole image and contains rich context information, enabling it to fit large scale facial components like the face contour. While PAF is more adept at fitting facial features due to the implicit frontalizion, which can well assist PNCC.

\subsection{Analysis of Feature Properties}
In Sec.~\ref{sec-feature-design}, we introduce three requirements of the input feature: feedback, convolvable and convergence. Among them, the benefits from convolvable and convergence may not be obvious and are further evaluated here. Corresponding to PNCC and PAF, we propose two alternative input features which miss these two properties respectively.

\begin{figure}[!htb]
  \centering
  \subfigure[PIndex]{\label{fig-convolvable-a}
  \includegraphics[width=0.225\textwidth]{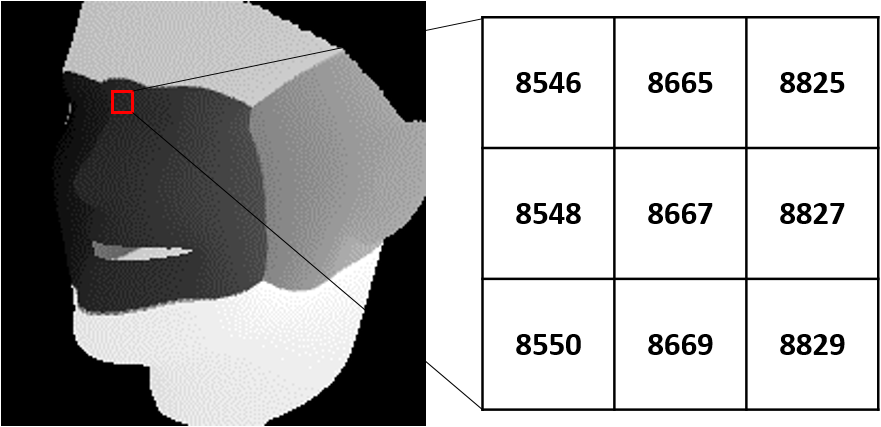}}
  \subfigure[PNCC]{\label{fig-convolvable-b}
  \includegraphics[width=0.225\textwidth]{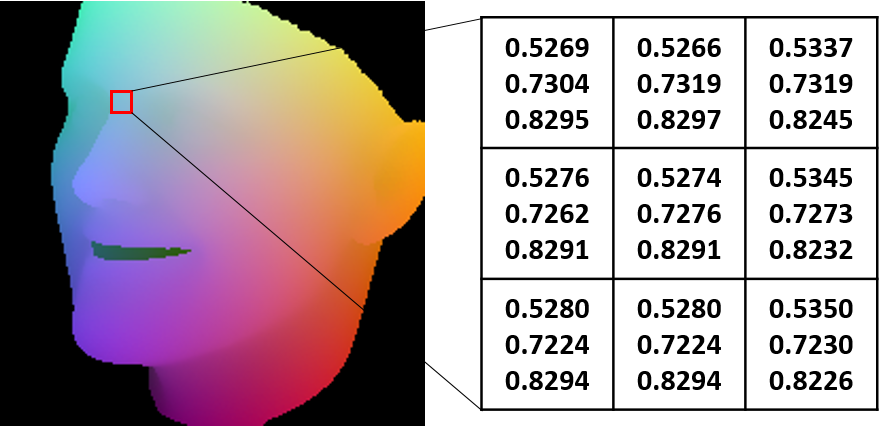}}
  \caption{The convolvable property of PNCC and PIndex: (a) A local patch of PIndex. The values can only be smooth in the indexing direction (vertical in this figure). (b) A local patch of PNCC. Values are smooth in 2D along each channel.}
  \label{fig-convolvable}
\end{figure}

\vspace{0.3em}
\noindent\textbf{Convovable Property}: As the alternative to PNCC, we propose the Projected Index (PIndex) which renders the projected 3D face with the $1$-channel vertex index (from $1$ to $53,490$ in BFM~\cite{Paysan-AVSS-09}) rather than the $3$-channel NCC, see Fig.~\ref{fig-convolvable}.
Note that even though PIndex provides the semantic meaning of each pixel, it is not smooth and the convolution of vertex indexes on a local patch is hard to be interpreted by the CNN. As a result, PIndex violates the convolvable requirement. Using the PNCC stream as the network, we adopt PNCC and PIndex as the input feature respectively. As shown in Table~\ref{tab-feature-baseline}, by violating the convolvable requirement, the performance drops since the learning task becomes more difficult.

\begin{table}[!htb]\small
 \tabcolsep 7pt \caption{The NME(\%) of PAF, PNCC and their corresponding alternative features, evaluated on AFLW2000-3D with different yaw interval.}
  \begin{center}
  \begin{tabular}{ c c  c  c  c  }
    \hline
    \multirow{2}{*}{Feature} & \multirow{2}{*}{$[0,30]$} & \multirow{2}{*}{$[30,60]$} & \multirow{2}{*}{$[60,90]$}  & \multirow{2}{*}{Mean}  \\
    &  & &  &     \\
    \hline
    \hline
    PIndex & 3.33 & 3.95 & 5.60 & 4.29  \\
    PNCC & 3.14 & 3.81 & 5.35 & 4.10  \\
    \hline
    TM & 3.38 & 4.48 & 6.76 & 4.87 \\
    PAF & 3.20 & 4.12 & 6.36 & 4.56  \\
    \hline
  \end{tabular}
  \end{center}
  \label{tab-feature-baseline}
\end{table}

\begin{figure}[!htb]
  \centering
  \subfigure[TM (Sub-Fitted)]{\label{fig-convergence-a}
  \includegraphics[width=0.1125\textwidth]{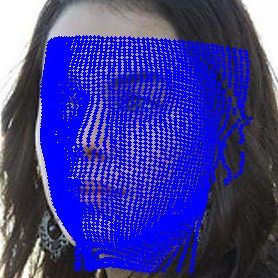}
  \includegraphics[width=0.1125\textwidth]{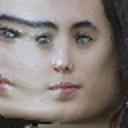}}
  \subfigure[TM (Fitted)]{\label{fig-convergence-b}
  \includegraphics[width=0.1125\textwidth]{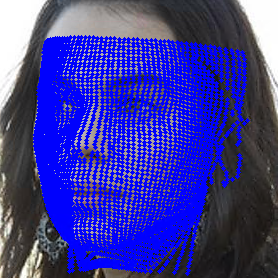}
  \includegraphics[width=0.1125\textwidth]{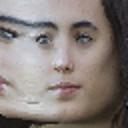}}\\
  \subfigure[PAF (Sub-Fitted)]{\label{fig-convergence-c}
  \includegraphics[width=0.1125\textwidth]{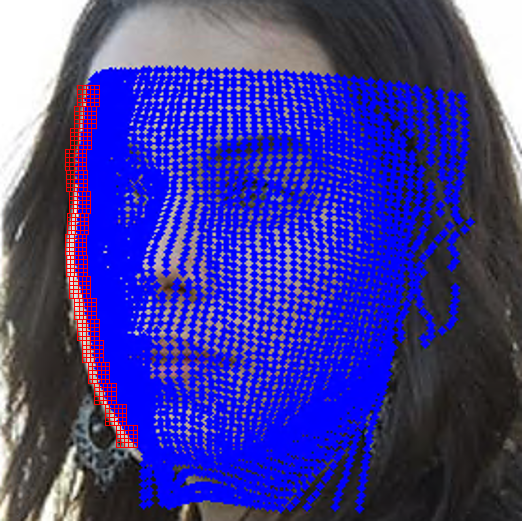}
  \includegraphics[width=0.1125\textwidth]{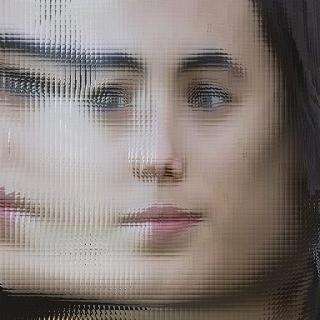}}
  \subfigure[PAF (Fitted)]{\label{fig-convergence-d}
  \includegraphics[width=0.1125\textwidth]{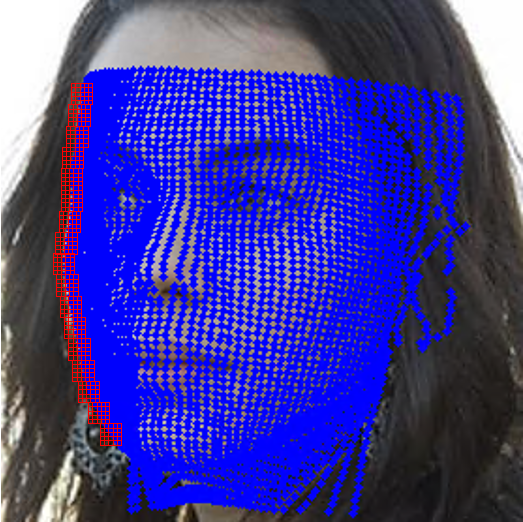}
  \includegraphics[width=0.1125\textwidth]{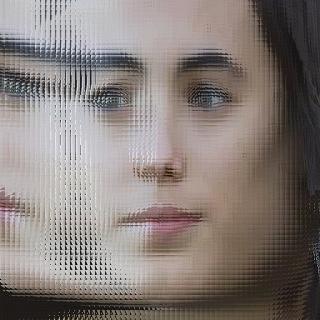}}\\
  \caption{The convergence property of TM and PAF. The first row: the mapped textures from a sub-fitted (a) and a fitted (b) sample. They show very similar appearances. The second row: the feature patch maps of PAF from a sub-fitted (c) and a fitted (d) sample. The convolution on the face contour vertices (the red grid) cover the pixels beyond the face region, enable PAF to exhibit discriminative appearance when the face contour is fitted.}
  \label{fig-convergence}
\end{figure}

\vspace{0.3em}
\noindent\textbf{Convergence Property}: As the alternative to PAF, we propose the Texture Mapping (TM)~\cite{Aldrian-PAMI-13} which rearranges the pixels on the projected feature anchors to a $64\times64$ image, see Fig.~\ref{fig-convergence}.
Compared with PAF, the main drawback of TM is the weak description beyond the model region. As shown in Fig.~\ref{fig-convergence-a} and Fig.~\ref{fig-convergence-b}, TM cannot discriminate whether the projected 3D model occludes the face in the image completely~\cite{piotraschke2016automated}. As a result, whether the fitting is complete is not discriminative for TM, which means the convergence requirement is not fulfilled. On the contrary, PAF can better describe the context information with the convolution on the face contour vertices.  As shown in Fig.~\ref{fig-convergence-c} and Fig.~\ref{fig-convergence-d}, PAF shows different appearances before and after the face contour is fitted. Table~\ref{tab-feature-baseline} shows the results of PAF and TM which use the PAF stream as the network. We can see that PAF outperforms TM by over $6\%$ which verifies the effectiveness of the convergence property.

\subsection{Analysis of Cost Function}
\begin{figure}[!htb]
  \centering
  \includegraphics[width=0.35\textwidth]{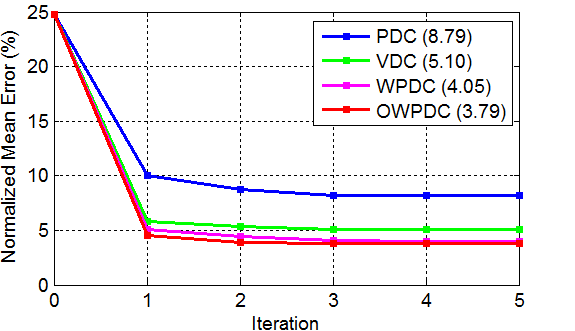}
  \caption{The testing errors with different cost functions, evaluated on AFLW2000-3D. The value in the bracket indicates the NME after the third iteration.}
  \label{fig-error-cost}
\end{figure}

\noindent\textbf{Performance with Different Cost}: We demonstrate the errors along the cascade with different cost functions including PDC, VDC, WPDC and OWPDC. Fig.~\ref{fig-error-cost} demonstrates the testing error at each iteration. All the networks are trained until convergence.
It is shown that PDC cannot well model the fitting error and converges to an unsatisfied result. VDC is better than PDC, but the pathological curvature problem makes it only concentrate on a small set of parameters and limits its performance. WPDC models the importance of each parameter and achieves a better result. Finally OWPDC further models the parameter priority, leading to faster convergence and the best performance.

\begin{figure}[!htb]
  \centering
  \includegraphics[width=0.45\textwidth]{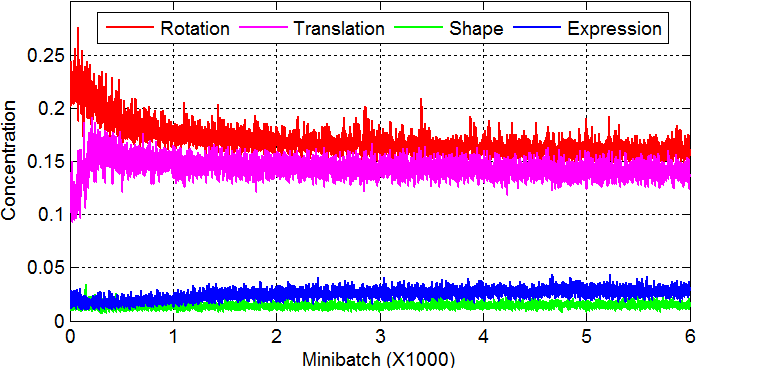}
  \caption{The mean weights of each mini-batch along the training process in the first iteration. The weights are normalized by $\textbf{w}/\sum\textbf{w}$ for better representation. The curves indicate the max value among the quaternion (rotation curve), $x$ and $y$ translation (translation curve), PCA shape (shape curve) and expression parameters (expression curve).}
  \label{fig-weights}
\end{figure}

\vspace{0.3em}
\noindent\textbf{Weights of OWPDC}: Since the weights of OWPDC reflect the priority of parameters, how the priority changes along the training process is also an interesting point to investigate. In this experiment, for each mini-batch during training, we record the mean weights of the mini-batch and plot the mini-batch weight in Fig.~\ref{fig-weights}. It can be seen that at beginning, the pose parameters (rotation and translation) show much higher priority than morphing parameters (shape and expression). As the training proceeds with error reducing, the pose weights begin to decrease and the CNN deals out its concentration to morphing parameters.

\subsection{Error Reduction in Cascade}
\begin{figure}[!htb]
  \centering
  \subfigure[]{
  \includegraphics[width=0.225\textwidth]{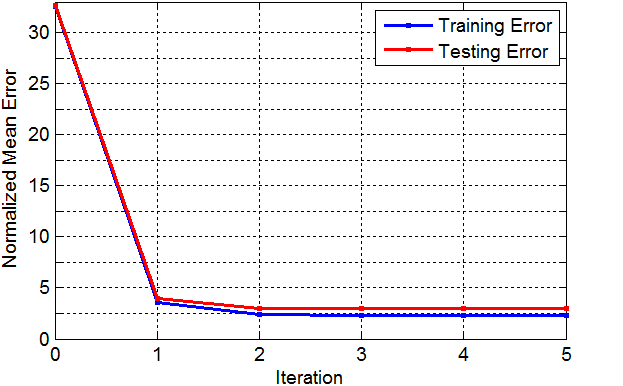}}
  \subfigure[]{
  \includegraphics[width=0.225\textwidth]{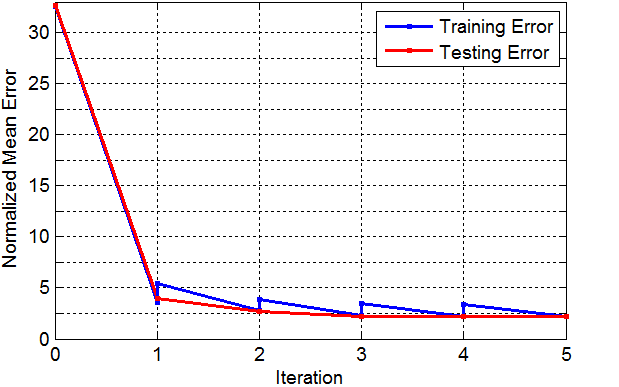}}
  \caption{The training and testing errors without (a) and with (b) initialization regeneration.}
  \label{fig-error-cascade}
\end{figure}

To analyze the overfitting problem in Cascaded Regression and evaluate the effectiveness of initialization regeneration, we divide 300W-LP into $97,967$ samples for training and $24,483$ samples for testing, without identity overlapping. Fig.~\ref{fig-error-cascade} shows the training and testing errors at each iteration, without and with initialization regeneration.
As observed, in traditional Cascaded Regression the training and testing errors converge fast after two iterations. While with initialization regeneration, the training error is updated at the beginning of each iteration and the testing error continues descending. Considering both effectiveness and efficiency we choose three iterations in 3DDFA.

\begin{figure}
  \centering
  \includegraphics[width=0.48\textwidth]{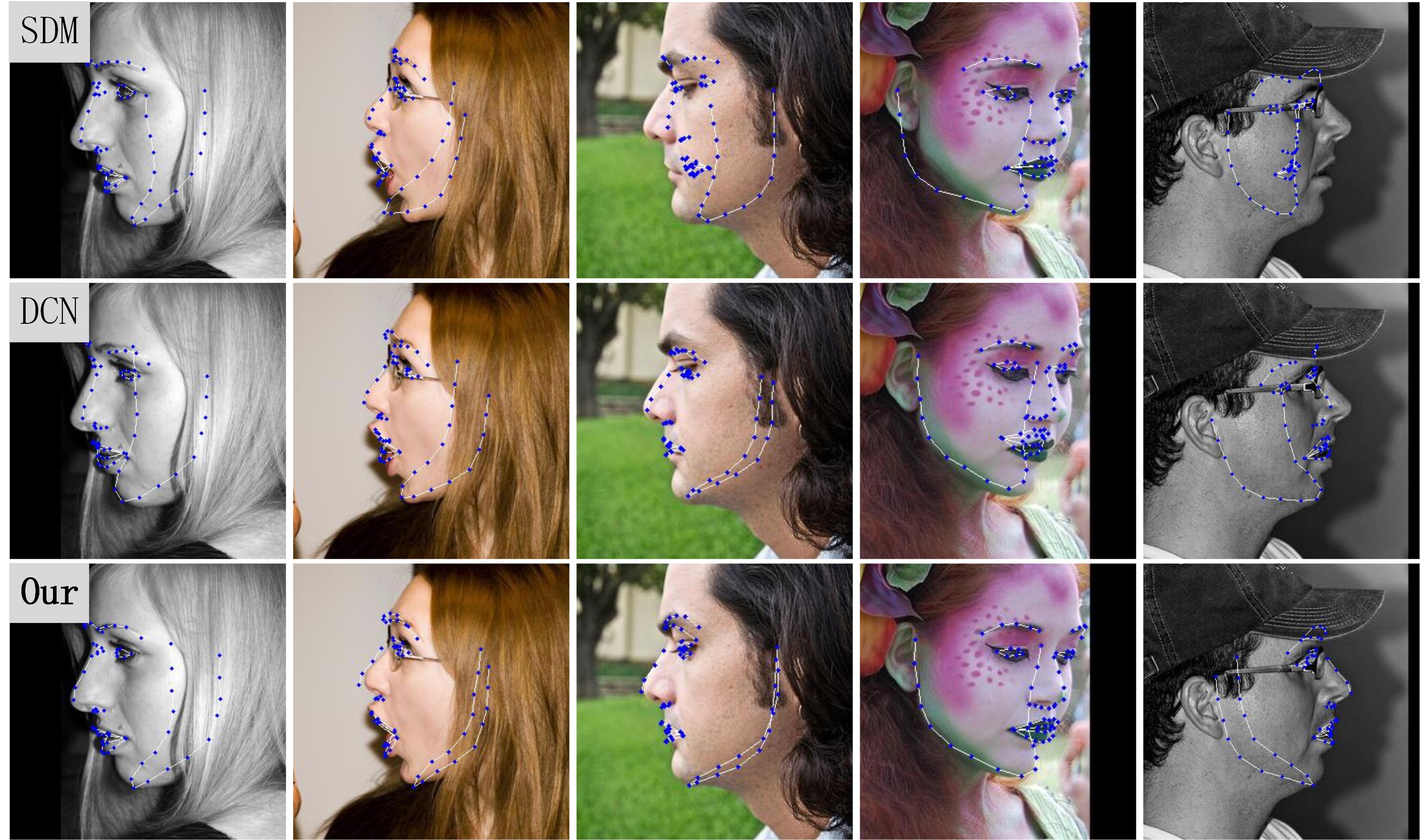}
  \caption{Results of SDM, DCN and our approach on AFLW.}
  \label{fig-demo-compare}
\end{figure}

\subsection{Comparison Experiments}
In this paper, we evaluate the performance of 3DDFA on three different tasks: the large-pose face alignment on AFLW, the 3D face alignment on AFLW2000-3D and the medium-pose face alignment on 300W.

\begin{table*}\small
 \tabcolsep 7pt \caption{The NME(\%) of face alignment results on AFLW and AFLW2000-3D with the first and the second best results highlighted. The brackets show the training sets.}
  \begin{center}
  \begin{tabular}{| c || c | c | c | c | c || c | c | c | c | c |}
    \hline
     &\multicolumn{5}{c||}{AFLW Dataset (21 pts)} & \multicolumn{5}{c|}{AFLW2000-3D Dataset (68 pts)}\\
    \hline
    \multirow{2}{*}{Method} & \multirow{2}{*}{$[0,30]$} & \multirow{2}{*}{$[30,60]$} & \multirow{2}{*}{$[60,90]$}  & \multirow{2}{*}{Mean} & \multirow{2}{*}{Std} & \multirow{2}{*}{$[0,30]$} & \multirow{2}{*}{$[30,60]$} & \multirow{2}{*}{$[60,90]$}  & \multirow{2}{*}{Mean} & \multirow{2}{*}{Std}\\
    &  & &  &  &  &  &  &  &  & \\
    \hline
    LBF(300W) & 7.17 & 17.54 & 28.45 & 17.72 & 10.64 & 6.17 & 16.48 & 25.90 & 16.19 & 9.87\\
    LBF(300W-LP) & 8.43 & 9.54 & 13.06 & 10.34 & 2.42 & 8.15 & 9.49 & 12.91 & 10.19 & 2.45\\
    \hline
    ESR(300W) & 5.58 & 10.62 & 20.02 & 12.07 & 7.33 & 4.38 & 10.47 & 20.31 & 11.72 & 8.04\\
    ESR(300W-LP) & 5.66 & 7.12 & 11.94 & 8.24 &  3.29 & 4.60 & 6.70 & 12.67 & 7.99 &  4.19\\
    \hline
    CFSS(300W) & 4.68 & 9.78 & 23.07 & 12.51 & 9.49 & \textbf{3.44} & 10.90 & 24.72 & 13.02 & 10.08\\
    CFSS(300W-LP) & 5.42 & 6.73 & 11.48 & 7.88 & 3.19 & 4.77 & 6.71 & 11.79 & 7.76 & 3.63\\
    \hline
    RCPR(300W) & 5.40 & 9.80 & 20.61 & 11.94 & 7.83 & 4.16 & 9.88 & 22.58 & 12.21 & 9.43\\
    RCPR(300W-LP) & 5.43 & 6.58 & 11.53 & 7.85 & 3.24 & 4.26 & 5.96 & 13.18 & 7.80 & 4.74\\
    \hline
    MDM(300W) & 5.14 & 10.95 & 24.11 & 13.40 & 9.72 & 4.64 & 10.35 & 24.21 & 13.07 & 10.07\\
    MDM(300W-LP) & 5.57 & 5.99 & 9.96 & 7.17 & 2.43 & 4.85 & 5.92 & 8.47 & 6.41 & 1.86\\
    \hline
    SDM(300W) & \textbf{4.67} & 6.78 & 16.13 & 9.19 & 6.10 & 3.56 & 7.08 & 17.48 & 9.37 & 7.23\\
    SDM(300W-LP) & 4.75 & 5.55 & 9.34 & 6.55 & 2.45 & 3.67 & 4.94 & 9.76 & 6.12 & 3.21\\
    \hline
    TSPM(300W-LP) & 5.91 & 6.52 & 7.68 & 6.70 & 0.90 & - & - & - & - & -\\
    \hline
    RMFA & 5.67 & 7.77 & 11.29 & 8.24 & 2.84 & 4.96 & 8.44 & 13.92 & 9.11 & 4.52 \\
    \hline
    DCN(300W-LP) & 4.99 & 5.47 & 8.10 & 6.19 & 1.68 & 3.93 & 4.67 & \textbf{7.71} & 5.44 & \textbf{2.00}\\
    \hline
    \hline
    \textbf{3DDFA(Pre)}~\cite{zhu2016face} & 5.00 & \textbf{5.06} & \textbf{6.74} & \textbf{5.60} & \textbf{0.99} & 3.78 & \textbf{4.54} & 7.93 & \textbf{5.42} & 2.21\\
    \textbf{Proposed} & \textbf{4.11} & \textbf{4.38} & \textbf{5.16} & \textbf{4.55} & \textbf{0.54} & \textbf{2.84} & \textbf{3.57} & \textbf{4.96} & \textbf{3.79} & \textbf{1.08}\\
    \hline
  \end{tabular}
  \end{center}
  \label{tab-falp}
\end{table*}

\subsubsection{Large Pose Face Alignment on AFLW}
\noindent\textbf{Protocol:}
In this experiment, we regard the whole AFLW as the testing set and divide it into three subsets according to their absolute yaw angles: $[0^{\circ},30^{\circ}]$, $[30^{\circ},60^{\circ}]$, and $[60^{\circ},90^{\circ}]$ with $11,596$, $5,457$ and $4,027$ samples respectively.
The alignment accuracy is evaluated by the Normalized Mean Error (NME), which is the average of landmarks error normalised by face size~\cite{jourabloo2015pose}. The face size is defined as the $\sqrt{width*height}$ of the bounding box (the rectangle hull of all the $68$ landmarks). Besides, we report the standard deviation of NMEs across testing subsets to measure the pose robustness. During training, we use the projected 3D landmarks as the ground truth to train 2D methods. For convenient comparison, the ground truth bounding boxes are used for initialization.

\vspace{0.3em}
\noindent\textbf{Methods:}
Since little experiment has been conducted on the whole AFLW, we choose some baselines with released training codes, including RCPR~\cite{burgos2013robust}, ESR~\cite{Cao-CVPR-12}, LBF~\cite{ren2014face}, CFSS~\cite{zhu2015face}, SDM~\cite{yan2013learn}, MDM~\cite{trigeorgis2016mnemonic}, RMFA~\cite{chen2016robust} and TSPM~\cite{zhu2012face}. Among them RCPR is an occlusion-robust method with the potential to deal with self-occlusion and we train it with landmark visibility computed by 3D information~\cite{hassner2014effective}. ESR, SDM, LBF and CFSS are popular Cascaded Regression based methods, among which SDM~\cite{yan2013learn} is the winner of ICCV2013 300W face alignment challenge. MDM is a deep learning base method which adopts CNNs to extract image features. TSPM and RMFA adopt the multi-view framework which can deal with large poses.
Besides the state-of-the-art methods, we introduce a Deep Convolutional Network (DCN) as a CNN based baseline. DCN directly regresses raw image pixels to the landmark positions with a CNN. The CNN has five convolutional layers, four pooling layers and two fully connected layers (the same as the PNCC stream) to estimate $68$ landmarks from a $200\times200\times3$ input image. Besides, we also compare with our previous work~\cite{zhu2016face} but we do not adopt the SDM based landmark refinement here.

\begin{figure*}[!htb]
  \centering
  \subfigure[$0^{\circ}$ to $30^{\circ}$]{
  \label{fig-ced-aflw-0-30}
  \includegraphics[width=0.225\textwidth]{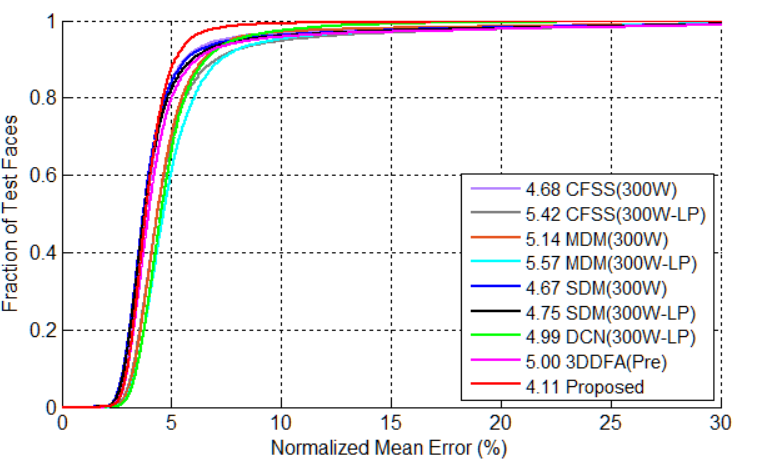}}
  \subfigure[$30^{\circ}$ to $60^{\circ}$]{
  \label{fig-ced-aflw-30-60}
  \includegraphics[width=0.225\textwidth]{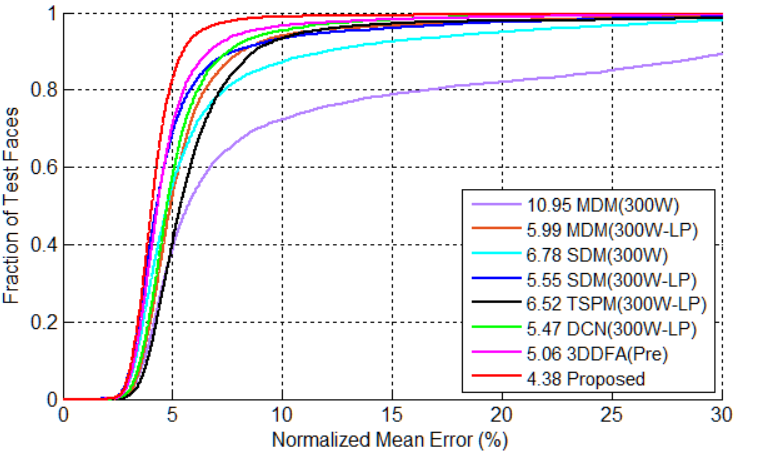}}
  \subfigure[$60^{\circ}$ to $90^{\circ}$]{
  \label{fig-ced-aflw-60-90}
  \includegraphics[width=0.225\textwidth]{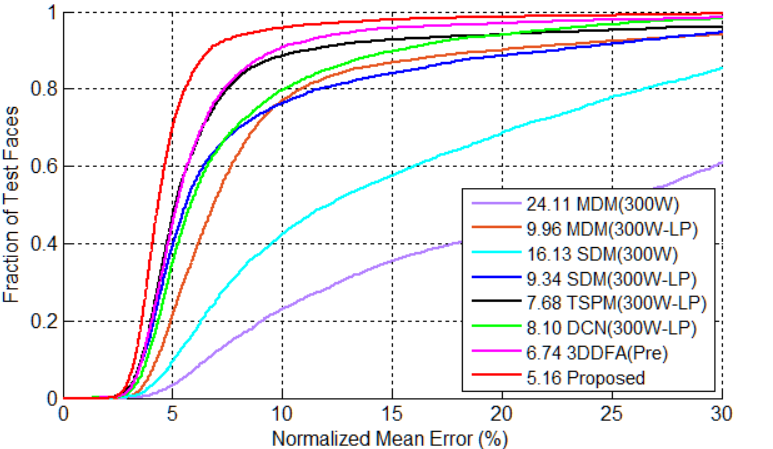}}
  \subfigure[Mean]{
  \label{fig-ced-aflw-mean}
  \includegraphics[width=0.225\textwidth]{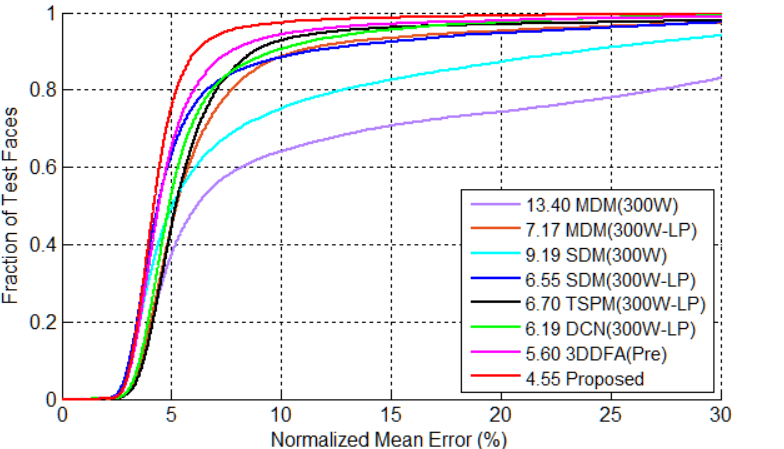}}
  \caption{Comparisons of cumulative errors distribution (CED) curves on AFLW with yaw distributing at: (a) $[0^{\circ},30^{\circ}]$, (b) $[30^{\circ},60^{\circ}]$ and (c) $[60^{\circ},90^{\circ}]$. We further plot a mean CED curve (d) with a subset of 12,081 samples whose absolute yaw angles within each yaw iterval are 1/3 each. Only the top $6$ methods are shown.}
  \label{fig-ced-aflw}
\end{figure*}

Table~\ref{tab-falp} shows the comparison results and Fig.~\ref{fig-ced-aflw} shows the corresponding CED curves. Each 2D method is trained on 300W and 300W-LP respectively to demonstrate the boost from face profiling. For DCN, 3DDFA and TSPM which depend on large scales of data or large-pose data, we only evaluate the models trained on 300W-LP. Given that RMFA only releases the testing code, we just evaluate it with the provided model. Besides, in large poses TSPM model only detects $10$ of the $21$ landmarks, we only evaluate the error of the $10$ points for TSPM.

\begin{figure*}[!htb]
  \centering
  \subfigure[$0^{\circ}$ to $30^{\circ}$]{
  \label{fig-ced-aflw2000-0-30}
  \includegraphics[width=0.225\textwidth]{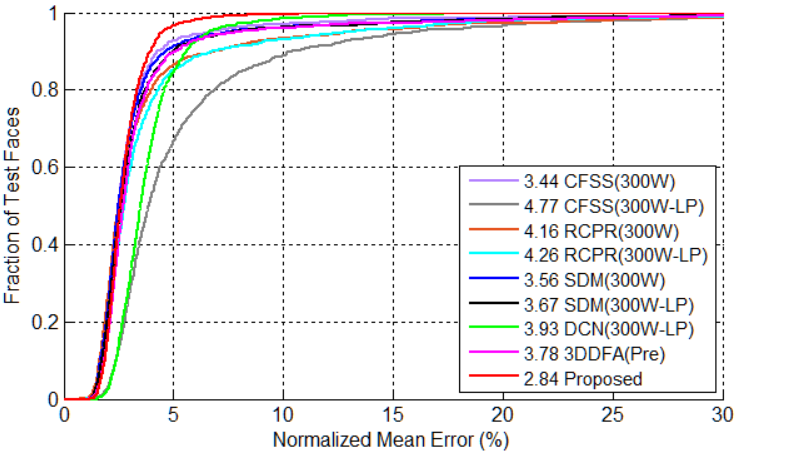}}
  \subfigure[$30^{\circ}$ to $60^{\circ}$]{
  \label{fig-ced-aflw2000-30-60}
  \includegraphics[width=0.225\textwidth]{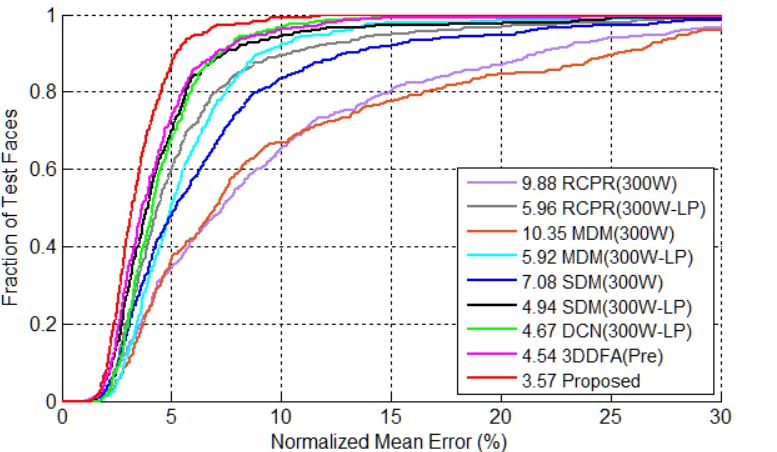}}
  \subfigure[$60^{\circ}$ to $90^{\circ}$]{
  \label{fig-ced-aflw2000-60-90}
  \includegraphics[width=0.225\textwidth]{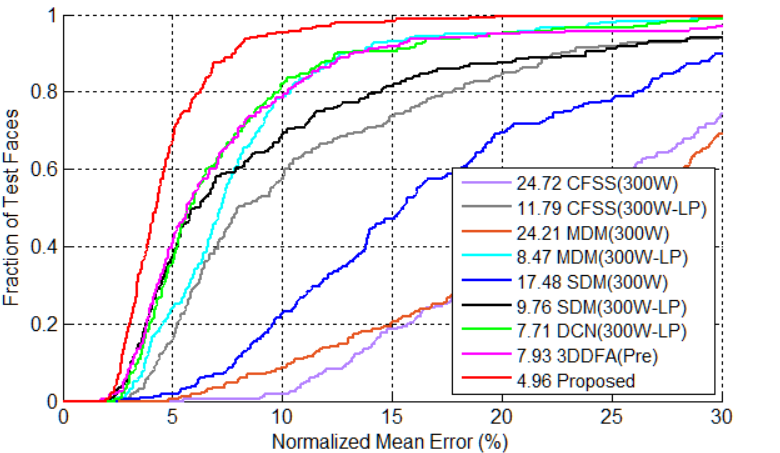}}
  \subfigure[Mean]{
  \label{fig-ced-aflw2000-mean}
  \includegraphics[width=0.23\textwidth]{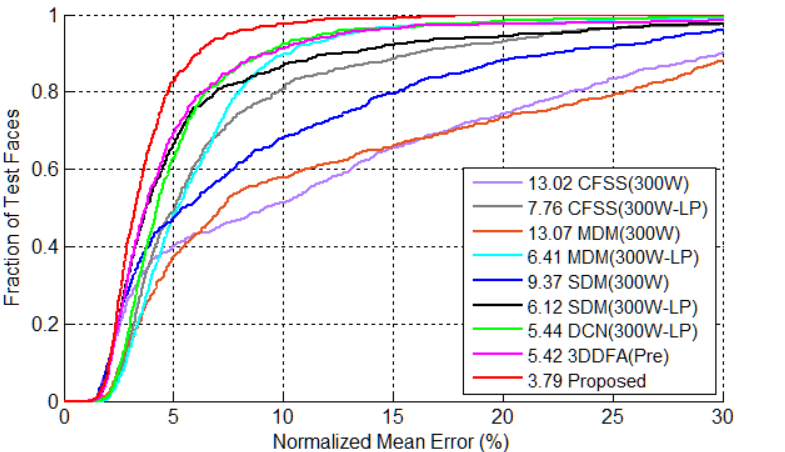}}
  \caption{Comparisons of cumulative errors distribution (CED) curves on AFLW2000-3D with yaw distributing at: (a) $[0^{\circ},30^{\circ}]$, (b) $[30^{\circ},60^{\circ}]$ and (c) $[60^{\circ},90^{\circ}]$. We further plot a mean CED curve (d) with a subset of 696 samples whose absolute yaw angles within each yaw iterval are 1/3 each. Only the top $6$ methods are shown.}
  \label{fig-ced-aflw2000}
\end{figure*}

\vspace{0.3em}
\noindent\textbf{Results:}
Firstly, the results indicate that all the methods benefit substantially from face profiling when dealing with large poses. The improvements in $[60^{\circ},90^{\circ}]$ exceed $40\%$ for all the methods. This is especially impressive since the alignment models are trained on the synthesized data and tested on real samples, which well demonstrates the fidelity of face profiling. Secondly, in near frontal view, most of methods show very similar performance as shown in Fig~\ref{fig-ced-aflw-0-30}. As the yaw angle increases in Fig~\ref{fig-ced-aflw-30-60} and Fig~\ref{fig-ced-aflw-60-90}, most of 2D methods begin to degrade but 3DDFA could still maintain its performance. Finally, 3DDFA reaches the state of the art above all the 2D methods especially beyond medium poses. The minimum standard deviation also demonstrates its robustness to pose variations.

\begin{figure}
  \centering
  \subfigure[]{\includegraphics[width=0.11\textwidth]{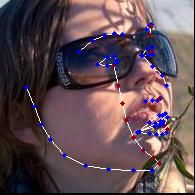}}
  \subfigure[]{\includegraphics[width=0.11\textwidth]{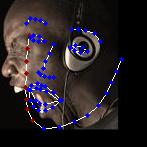}}
  \subfigure[]{\includegraphics[width=0.11\textwidth]{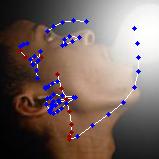}}
  \subfigure[]{\includegraphics[width=0.11\textwidth]{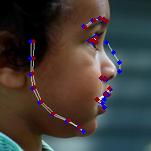}}
  \caption{Typical failure reasons of 3DDFA, including (a) complicated shadow and occlusion, (b) extreme pose and expression, (c) extreme illumination and (d) limited shape variations of 3DMM on nose.}
  \label{fig-demo-failure}
\end{figure}

In Fig.~\ref{fig-demo-compare}, we demonstrate some alignment results of 3DDFA and representative 2D methods. Besides, Fig.~\ref{fig-demo-failure} show some typical failure cases.

\subsubsection{3D Face Alignment in AFLW2000-3D}
As described in Section~\ref{sec-datasets}, 3D face alignment evaluation can be degraded to full-landmarks evaluation considering both visible and invisible ones. Using AFLW2000-3D as the testing set, this experiment follows the same protocol as AFLW, except all the $68$ landmarks are used for evaluation. There are $1,306$ samples in $[0^{\circ},30^{\circ}]$, $462$ samples in $[30^{\circ},60^{\circ}]$ and $232$ samples in $[60^{\circ},90^{\circ}]$. The results are demonstrated in Table~\ref{tab-falp} and the CED curves are ploted in Fig.~\ref{fig-ced-aflw2000}. We do not report the performance of TSPM models since they do not detect invisible landmarks.

Compared with the results in AFLW, we can see that the standard deviation is dramatically increased, meaning that it is more difficult to keep pose robustness when considering all the landmarks. Besides, the improvement of 3DDFA over the best 2D method DCN is increased from $26.49\%$ in AFLW to $30.33\%$ in AFLW2000-3D, which demonstrates the superiority of 3DDFA in 3D face alignment.

\subsubsection{Medium Pose Face Alignment}
As a face alignment approach to deal with full pose range, 3DDFA also shows competitive performance on the medium-pose 300W database, using the common protocol in~\cite{zhu2015face}. The alignment accuracy is evaluated by the standard landmark mean error normalized by the inter-pupil distance (NME). For 3DDFA, we sample the 68 landmarks from the fitted 3D face and refine them with SDM to reduce the labelling bias.
\begin{table}\small
 \tabcolsep 8pt \caption{The NME(\%) of face alignment results on 300W, with the first and the
second best results highlighted.}
  \begin{center}
  \begin{tabular}{ c  c  c  c }
    \hline
    Method & Common & Challenging & Full \\
    \hline
    TSPM~\cite{zhu2012face} & 8.22 & 18.33 & 10.20\\

    ESR~\cite{Cao-CVPR-12} & 5.28 & 17.00 & 7.58  \\

    RCPR~\cite{burgos2013robust} & 6.18 & 17.26 & 8.35 \\

    SDM~\cite{Xiong-CVPR-13} & 5.57 & 15.40 & 7.50 \\

    LBF~\cite{ren2014face} & 4.95 & 11.98 & 6.32 \\

    CFSS~\cite{zhu2015face} & \textbf{4.73} & 9.98 & 5.76 \\

    TCDCN~\cite{zhang2016learning} & \textbf{4.80} & \textbf{8.60} & \textbf{5.54} \\
    \hline
    3DDFA(Pre) & 5.53 & 9.56 & 6.31\\
    Proposed & 5.09 & \textbf{8.07} & \textbf{5.63}\\
    \hline
  \end{tabular}
  \end{center}
  \label{tab-fa-medium}
\end{table}
Table~\ref{tab-fa-medium} shows that even in medium poses 3DDFA performs competitively, especially on the challenging set.

\subsubsection{Robustness to Initialization}
The alignment performance can be greatly affected by the bounding boxes used for initialization. In this experiment, we initialize alignment methods with detected bounding boxes by FTF face detector~\cite{peiyun2017finding} rather than the ground truth bounding boxes. We drop the bad boxes whose IOU with ground truth bounding boxes are less than $0.6$ and generate the bounding boxes of undetected faces by random perturbation used in training. Table~\ref{tab-robust-init} shows the comparison results with the best two competitors DCN and SDM. Firstly, it can be seen that our method still outperforms others when initialized with face detectors. Besides, by comparing the performance drop brought by replacing bounding boxes, our method demonstrates best robustness to initialization.

\begin{table}\small
 \tabcolsep 4pt \caption{Alignment performance (NME) initialized by detected bounding boxes. The value in the brackets are the NME difference between results initialized by the detected and the ground truth bounding boxes}
  \begin{center}
  \begin{tabular}{| c || c | c | c || c | c | c |}
    \hline
     &\multicolumn{3}{c||}{AFLW} & \multicolumn{3}{c|}{AFLW2000-3D}\\
    \hline
              & SDM   & DCN  & Ours & SDM & DCN & Ours\\
    \hline
    $[0,30]$  & \tabincell{c}{5.09\\(0.34)}  & \tabincell{c}{5.31\\(0.32)} & \tabincell{c}{4.24\\(0.13)} & \tabincell{c}{4.11\\(0.44)}  & \tabincell{c}{4.34\\(0.41)} & \tabincell{c}{3.00\\(0.16)} \\
    \hline
    $[30,60]$ & \tabincell{c}{6.02\\(0.47)}  & \tabincell{c}{5.95\\(0.48)} & \tabincell{c}{4.59\\(0.21)} & \tabincell{c}{6.19\\(1.25)}  & \tabincell{c}{5.42\\(0.75)} & \tabincell{c}{3.89\\(0.32)} \\
    \hline
    $[60,90]$ & \tabincell{c}{10.13\\(0.79)}  & \tabincell{c}{8.13\\(0.03)} & \tabincell{c}{5.32\\(0.16)} & \tabincell{c}{12.03\\(2.27)}  & \tabincell{c}{8.72\\(1.01)} & \tabincell{c}{5.55\\(0.59)} \\
    \hline
    Mean      & \tabincell{c}{7.08\\(0.53)}  & \tabincell{c}{6.47\\(0.28)} & \tabincell{c}{4.72\\(0.17)} & \tabincell{c}{7.44\\(1.32)}  & \tabincell{c}{6.16\\(0.74)} & \tabincell{c}{4.15\\(0.36)} \\
    \hline
  \end{tabular}
  \end{center}
  \label{tab-robust-init}
\end{table}

\section{Conclusions}
Most of face alignment methods tend to fail in profile view since the self-occluded landmarks cannot be detected. Instead of the traditional landmark detection framework, this paper fits a dense 3D Morphable Model to achieve pose-free face alignment. By proposing two input features of PNCC and PAF, we cascade a couple of CNNs as a strong regressor to estimate model parameters. A novel OWPDC cost function is also proposed to consider the priority of parameters. To provide abundant samples for training, we propose a face profiling method to synthesize face appearances in profile views. Experiments show the state-of-the-art performance on AFLW, AFLW2000-3D and 300W.

\section{ACKNOWLEDGMENTS}
This work was supported by the National Key Research and Development Plan (Grant No.2016YFC0801002), the Chinese National Natural Science Foundation Projects \#61473291, \#61572501, \#61502491, \#61572536 and AuthenMetric R\&D Funds. Zhen Lei is the corresponding author.

\small
\bibliographystyle{IEEEtran}
\bibliography{reference_zxy}

\begin{IEEEbiography}[{\includegraphics[width=1in,height=1.25in,clip,keepaspectratio]{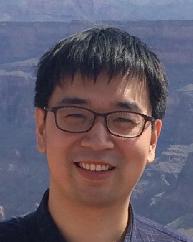}}]{Xiangyu Zhu}
received the BS degree in Sichuan University (SCU) in 2012, and the PhD degree from Institute of Automation, Chinese Academy of Sciences, in 2017, where he is currently an assistant professor. His research interests include pattern recognition and computer vision, in particular, image processing, 3D face model, face alignment and face recognition.
\end{IEEEbiography}
\vspace{-10mm}

\begin{IEEEbiography}[{\includegraphics[width=1in,height=1.25in,clip,keepaspectratio]{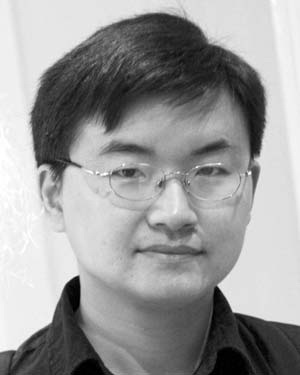}}]{Xiaoming Liu}
is an Assistant Professor at the Department of Computer Science and Engineering of Michigan State University. He received the Ph.D. degree in Electrical and Computer Engineering from Carnegie Mellon
University in 2004. Before joining MSU in Fall 2012, he was a research scientist at General Electric Global Research. His main research areas are human face recognition, biometrics, human computer interface, object tracking/recognition, online learning, computer vision, and pattern recognition.
\end{IEEEbiography}
\vspace{-10mm}

\begin{IEEEbiography}[{\includegraphics[width=1in,height=1.25in,clip,keepaspectratio]{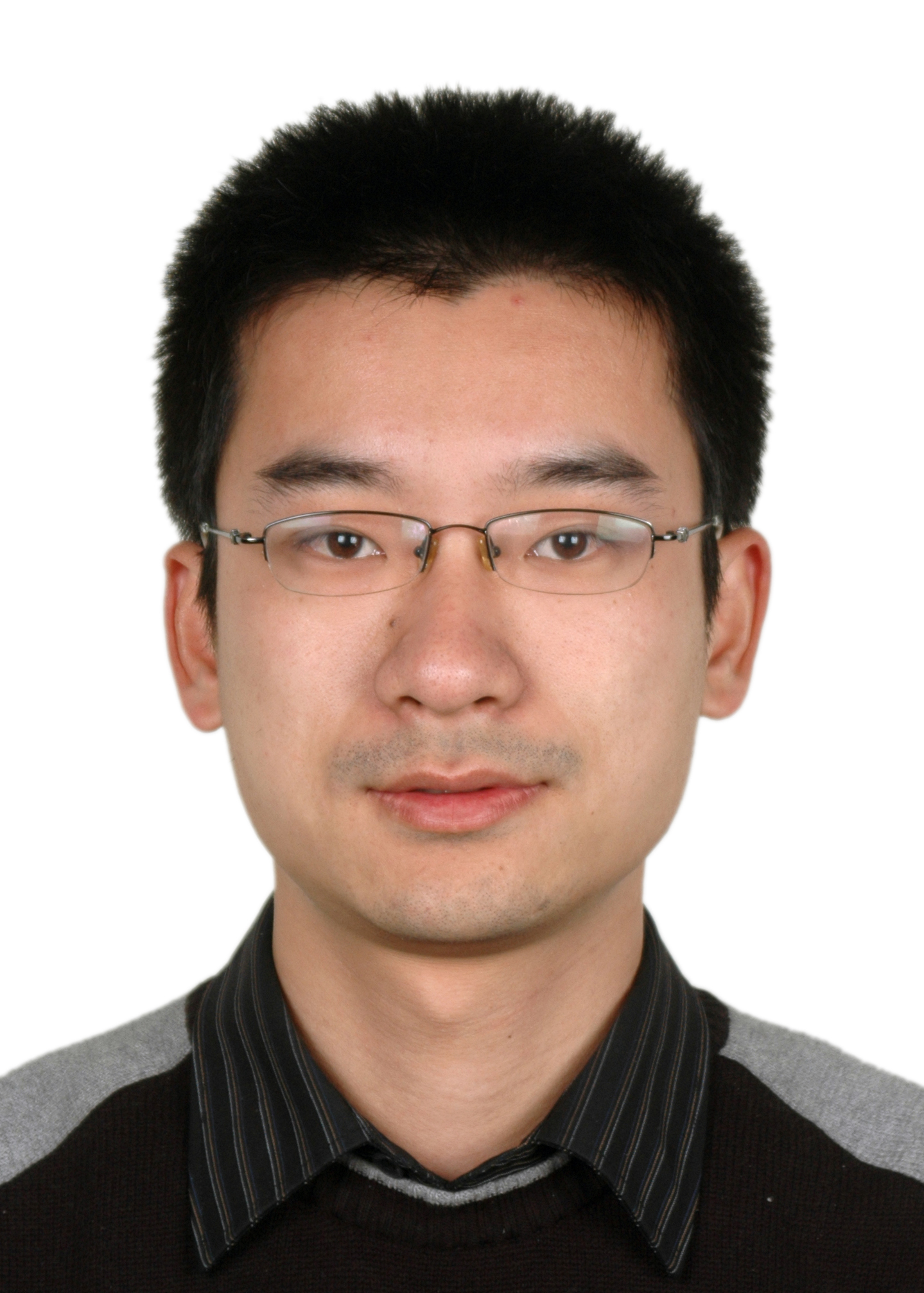}}]{Zhen Lei}
received the BS degree in automation from the University of Science and Technology of China, in 2005, and the PhD degree from the Institute of Automation, Chinese Academy of Sciences,in 2010, where he is currently an associate professor. His research interests are in computer vision, pattern recognition, image processing, and face recognition in particular.
\end{IEEEbiography}
\vspace{-10mm}

\begin{IEEEbiography}[{\includegraphics[width=1in,height=1.25in,clip,keepaspectratio]{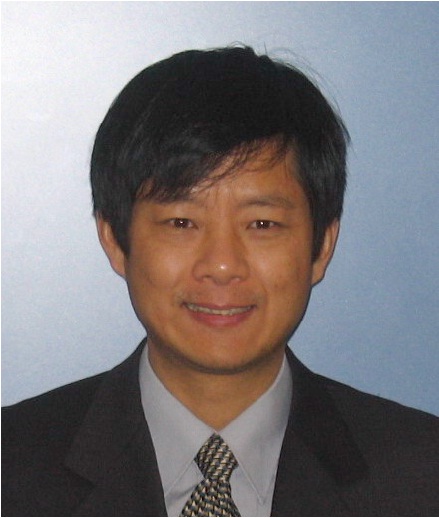}}]{Stan Z. Li}
received his B.Eng from Hunan University, China, M.Eng from National University of Defense Technology, China, and PhD degree from Surrey University, UK. He is currently a professor and the director of Center for Biometrics and Security Research (CBSR), Institute of Automation, Chinese Academy of Sciences (CASIA). He worked at Microsoft Research Asia as a researcher from 2000 to 2004. Prior to that, he was an associate professor at Nanyang Technological University, Singapore. He was elevated to IEEE Fellow for his contributions to the fields of face recognition, pattern recognition and computer vision. His research interest includes pattern recognition and machine learning, image and vision processing, face recognition, biometrics, and intelligent video surveillance.
\end{IEEEbiography}

\end{document}